\documentclass[11pt]{article}

\usepackage[margin=1in]{geometry}
\usepackage{times}
\usepackage[T1]{fontenc}
\usepackage[utf8]{inputenc}
\usepackage{amsmath,amssymb}
\usepackage{pifont}
\newcommand{\cmark}{\ding{51}}
\newcommand{\xmark}{\ding{55}}
\usepackage{graphicx}
\usepackage{booktabs}
\usepackage{multirow}
\usepackage{subcaption}
\usepackage{enumitem}
\usepackage[round]{natbib}
\usepackage{url}
\usepackage[colorlinks=true,linkcolor=blue,citecolor=blue,urlcolor=blue]{hyperref}

\usepackage{amsmath,amsfonts,bm}

\def\eqref#1{equation~\ref{#1}}

\def\1{\bm{1}}

\DeclareMathAlphabet{\mathsfit}{\encodingdefault}{\sfdefault}{m}{sl}
\SetMathAlphabet{\mathsfit}{bold}{\encodingdefault}{\sfdefault}{bx}{n}

\title{\textsc{CaptchaArena}: A Large-Scale, Fine-Grained Dataset for Training
Computer-Use Agents on Interactive CAPTCHAs}

\author{
  Zhenhao Zhang$^{1,\ast}$ \quad
  Zhaoyu Fan$^{2}$ \quad
  Haohan Ying$^{3}$ \quad
  Jingwen Hu$^{3}$ \quad
  Hancen Fan$^{1}$  \\[2pt]
  Junhao Zhou$^{4}$ \quad
  Zitian Chen$^{1}$ \quad
  Linchao Zhu$^{2,\dagger}$ \\[4pt]
  \normalsize $^{1}$Columbia University \quad
  $^{2}$Zhejiang University \\
  \normalsize $^{3}$University of Rochester \quad
  $^{4}$University of Illinois at Urbana-Champaign \\[3pt]
  \normalsize $^{\ast}$Project lead \quad $^{\dagger}$Corresponding author
}

\date{}

\begin{document}
\maketitle

\begin{abstract}
Interactive CAPTCHAs remain challenging for computer-use agents, while existing
datasets face trade-offs among type coverage, interaction fidelity, and trajectory supervision. To address these gaps, we present \textbf{CaptchaArena}, the first large-scale, fine-grained training dataset for interactive CAPTCHA solving. It contains $50$K puzzles across $20$ CAPTCHA types and $5$ interaction modes, with every solution verified through execution. \textbf{CaptchaArena} provides $50$K screenshot-action trajectories, including $46$K with step-by-step reasoning annotations. It also includes fine-grained pixel-mask annotations for irregular targets. Using \textbf{CaptchaArena}, we train \textbf{CaptchaAgent}, a single $9$B policy for all $20$ CAPTCHA types, with supervised fine-tuning followed by reinforcement learning. The environment verifier directly provides the RL reward. Supervised fine-tuning reaches $70.5$ Pass@1, and reinforcement learning further improves it to $71.7$, while also improving performance on two external benchmarks. These results demonstrate the value of large-scale, fine-grained computer-use supervision for training interactive CAPTCHA agents. We release \textbf{CaptchaArena} and \textbf{CaptchaAgent} at \url{https://github.com/X0X0X00/CaptchaArena}.
\end{abstract}

\section{Introduction}
\label{sec:intro}

Interactive CAPTCHAs expose a critical gap in current computer-use agents.
These agents can now perform long-horizon tasks using screenshots and
low-level mouse and keyboard actions
\citep{xie2024osworld}. However, a single failed CAPTCHA can stop an entire workflow and block everything downstream. Existing benchmarks often avoid measuring this, and each treats CAPTCHAs differently. Online-Mind2Web
\citep{xue2025onlinemind2web} removes CAPTCHA-protected websites during
construction. VisualWebArena \citep{koh2024visualwebarena} self-hosts every
site. Yet CAPTCHAs are worth evaluating because solving them requires the core loop of computer use. An agent must
read the web page, decide what to do, act on the right pixels, and verify
the resulting state change.

\begin{figure}[!t]
  \centering
  \includegraphics[width=\textwidth]{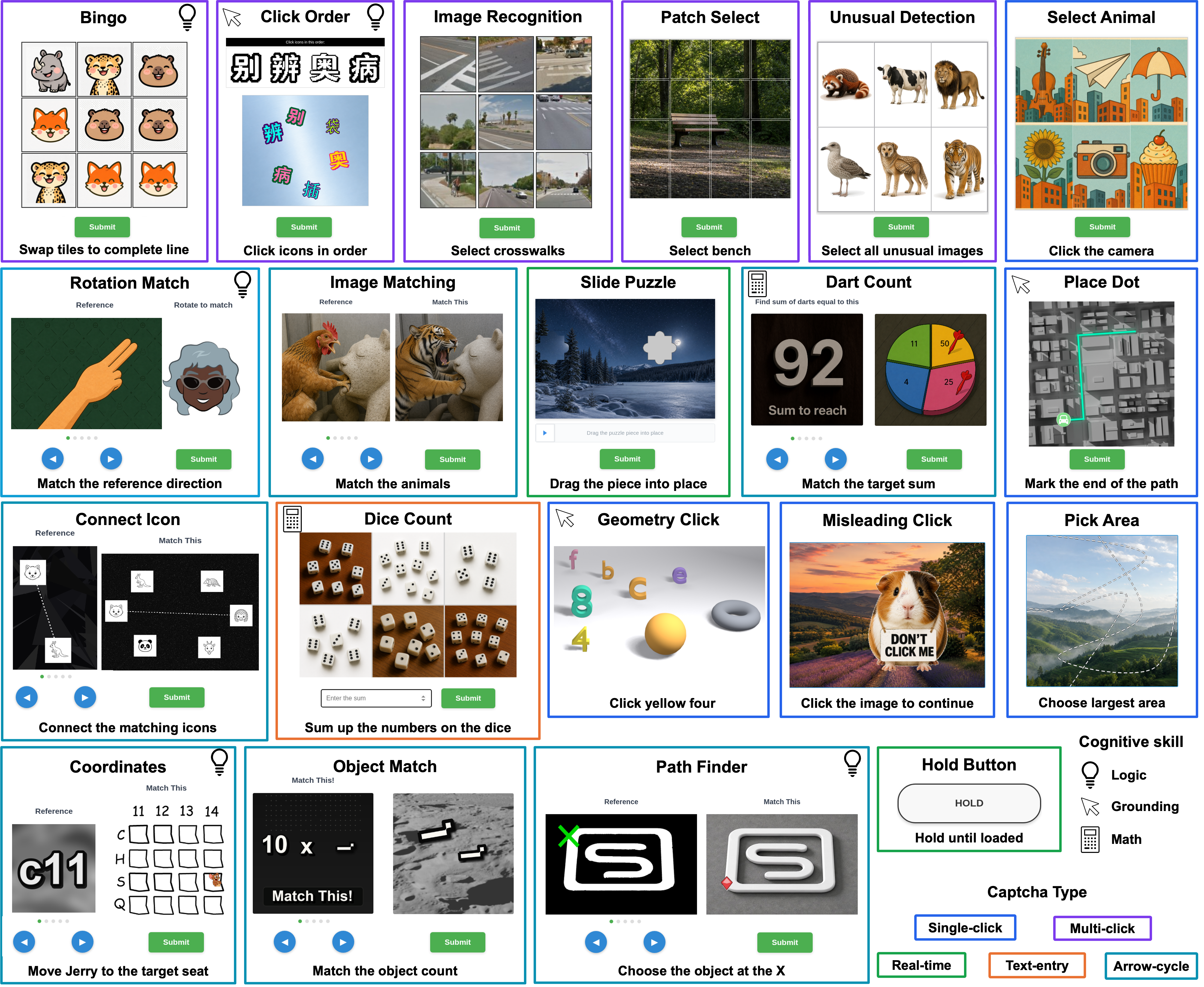}
  \caption{Representative puzzles from the $20$ CAPTCHA types in
  \textbf{CaptchaArena}, cropped for display. Border colors mark interaction
  modes and corner icons mark cognitive skills.}
  \label{fig:teaser}
\end{figure}

Training data for CAPTCHA interaction remain limited.
Existing CAPTCHA resources involve trade-offs among type coverage, interaction fidelity, and trajectory supervision. ReCAP \citep{chen2026recap} and
CaptchaMind \citep{wang2026captchamind} provide step-by-step interaction
trajectories with reasoning, but cover $7$ and $8$ challenge types. MirrorCAPTCHA
\citep{wu2026mirrorcaptcha} evaluates $18$ types, but its $370$K synthetic training samples provide only answer and click-coordinate supervision. Open CaptchaWorld
\citep{luo2025opencaptchaworld} covers $20$ types but evaluates agents through the browser-use framework. This framework cannot handle some CAPTCHA types that require fine-grained interaction. For example, on MirrorCAPTCHA, every browser-use agent scores $0$ on $6$ of its $18$ types \citep{wu2026mirrorcaptcha}. MCA-Bench
\citep{wu2026mcabench} predicts structured solutions and action traces rather than executing them step-by-step in a live browser environment. Its predicted clicks are graded against bounding boxes, while CAPTCHA-X \citep{song2025reasoningvision} uses manually defined geometric acceptance regions. Such region-based grading can over-accept areas outside irregular targets.

We introduce \textbf{CaptchaArena} to close these gaps
(Fig.~\ref{fig:teaser}). It is a large-scale, fine-grained computer-use training dataset containing $50$K interactive puzzles across $20$ types and $5$ interaction modes, including single-click, multi-click, arrow-cycle, real-time, and text-entry. Puzzle quality is manually inspected. For irregular targets, clicks are graded against pixel-masks. Each puzzle has an executable reference solution that is run in a real browser and must pass the environment verifier. The same execution produces one screenshot-action trajectory per puzzle. A teacher model adds step-by-step reasoning-annotations and a separate judge checks each annotation, producing $46$K reasoning-annotated trajectories.

We release \textbf{CaptchaAgent}, a single policy built on Qwen3.5-9B \citep{qwen35}. It is fine-tuned on a subset of the reasoning-annotated trajectories and then further trained with reinforcement
learning. The same environment verifier used to validate puzzle solutions directly provides the RL reward. Supervised fine-tuning raises Pass@1 from $11.4$ for the untrained backbone to $70.5$. Reinforcement learning further raises Pass@1 to $71.7$, and improves the score on two external benchmarks (App.~\ref{app:ocw}). Humans reach $94.1$ on the same $4{,}000$ puzzles through the same interface.

\paragraph{Contributions.}
\begin{itemize}[leftmargin=*,itemsep=0pt,parsep=0pt]
  \item \textbf{CaptchaArena}. A large-scale, fine-grained computer-use training dataset containing $50$K interactive CAPTCHA puzzles across $20$ types and $5$ interaction modes. Every solution is verified through execution, and the dataset includes $46$K reasoning-annotated trajectories and pixel-mask supervision.
  \item \textbf{CaptchaAgent}. A single $9$B policy covering all
    $20$ types, fine-tuned on a subset of the reasoning-annotated trajectories and then trained with
    reinforcement learning. The environment verifier provides the RL reward.
  \item \textbf{Error analysis}. A per-type analysis of remaining failures, comparing \textbf{CaptchaAgent} with the untrained backbone and human performance on the same puzzles.
\end{itemize}

\section{Related Work}
\label{sec:related}
\paragraph{Type coverage.}
Existing CAPTCHA benchmarks vary widely in type coverage (Table~\ref{tab:related}).
ReCAP \citep{chen2026recap} and CaptchaMind \citep{wang2026captchamind} provide training data for $7$ and $8$ CAPTCHA types. MCA-Bench \citep{wu2026mcabench} offers over $180$K training data for $20$ types, but trains task-specific LoRA adapters rather than a unified policy. MirrorCAPTCHA \citep{wu2026mirrorcaptcha} evaluates $1{,}000$ puzzles across $18$ types, including $6$ OCR variants. However, its $370$K synthetic training samples provide only answer and click-coordinate supervision. Open CaptchaWorld \citep{luo2025opencaptchaworld} covers $20$ types. Its current release contains $463$ puzzles, versus $225$ in the original paper. We adopt its task types. CAPTURE \citep{zhang2025capture} and Halligan \citep{teoh2025halligan}  evaluate $25$ and $26$ CAPTCHA types. Oedipus \citep{deng2025oedipus} evaluates only $4$ CAPTCHA types.

\paragraph{Interaction fidelity.}
Computer-use evaluation places an agent in a closed-loop perception-action process \citep{zhou2024webarena,xie2024osworld,xue2025onlinemind2web}. However, mainstream benchmarks generally do not evaluate CAPTCHA solving. \citet{akkil2026emergence} treat CAPTCHAs as external constraints during live evaluation, while HLL \citep{song2026hll} identifies CAPTCHA verification as a deployment bottleneck and finds frontier agents remain brittle. CAPTCHA-specific benchmarks also differ in their interaction interfaces. Open CaptchaWorld evaluates agents with the browser-use framework, whose actions often target indexed web elements rather than raw screen coordinates. MCA-Bench \citep{wu2026mcabench} produces structured outputs rather than step-by-step execution in a live browser.

\paragraph{Trajectory supervision.}
Existing resources also differ in the supervision they provide. CAPTCHA-X \citep{song2025reasoningvision} provides reasoning and action annotations for real-world CAPTCHA puzzles. CaptchaMind \citep{wang2026captchamind} provides process-level annotations across $8$ types and trains a unified policy with SFT followed by GRPO. ReCAP \citep{chen2026recap} collects large-scale reasoning-action and self-correction trajectories across $7$ types. Although ReCAP improves zero-shot transfer overall, its trained models still underperform their corresponding base models on some unseen CAPTCHA types. MirrorCAPTCHA \citep{wu2026mirrorcaptcha} instead trains a unified agent with a synthetic data pipeline, while MCA-Bench \citep{wu2026mcabench} fine-tunes task-specific LoRA adapters. Neither provides reasoning-annotated screenshot-action trajectories across broad CAPTCHA type coverage.

\begin{table}[t]
\centering
\footnotesize
\setlength{\tabcolsep}{5pt}
\renewcommand{\arraystretch}{1.1}
\caption{Comparison of CAPTCHA benchmarks and training datasets, as reported by each paper.}
\label{tab:related}
\begin{tabular}{@{}lcccccll@{}}
\toprule
Benchmark & Types & CU & Traj. & Reasoning & RL reward &
Click Eval. & \multicolumn{1}{c}{Scale} \\
\midrule
\multicolumn{8}{@{}l}{\emph{Evaluation-only benchmarks}} \\
Open CaptchaWorld & $20$ & \xmark & \xmark & \xmark & --     & --    &  $463^{\dagger}$ \\
Halligan          & $26$ & \cmark & \xmark & \xmark & --     & tol.  & $2{,}600$ \\
Oedipus           & $4$  & \xmark & \xmark & \xmark & --     & --    & $400$ \\
HLL               & $10$ & \cmark & \xmark & \xmark & --     & tol.  & -- \\
CAPTCHA-X         & $7$  & \cmark & \xmark & \cmark & --     & region   & $1{,}839$ \\
CAPTURE           & $25$ & \xmark & \xmark & \xmark & --     & --    & $61{,}464$ \\
\addlinespace
\multicolumn{8}{@{}l}{\emph{Training datasets}} \\
MCA-Bench         & $20$ & \xmark & \xmark & \xmark & \xmark & box & $180$K \\
MirrorCAPTCHA     & $18$ & \cmark & \xmark & \xmark & \xmark & --  & $370$K \\
ReCAP             & $7$  & \cmark & \cmark & \cmark & \xmark & --  & $160$K \\
CaptchaMind & $8$ & \cmark & \cmark & \cmark & \cmark & box & $16$K \\
\midrule
\textbf{CaptchaArena (ours)}
   & $\mathbf{20}$ & \cmark & \cmark & \cmark & \cmark & \textbf{tol.+mask}
   & $\mathbf{42}$\textbf{K} \\
\bottomrule
\end{tabular}

\vspace{3pt}
{\scriptsize
Scale denotes evaluation size for evaluation-only benchmarks and training size for training datasets.
\emph{CU} denotes direct low-level computer use rather than structured
prediction or browser-use/SoM \citep{yang2023setofmark, browser_use2024}.
\emph{Traj.} denotes multi-step screenshot-action trajectories with
intermediate observations.
Halligan and HLL report positional tolerance (\emph{tol.}) without specifying
its geometry or threshold.
A dash denotes not applicable or not reported.
$^{\dagger}$\,Later release; the paper reports $225$.\par}
\end{table}

\section{CaptchaArena}
\label{sec:data}
\begin{figure}[t]
  \centering
  \includegraphics[width=\textwidth]{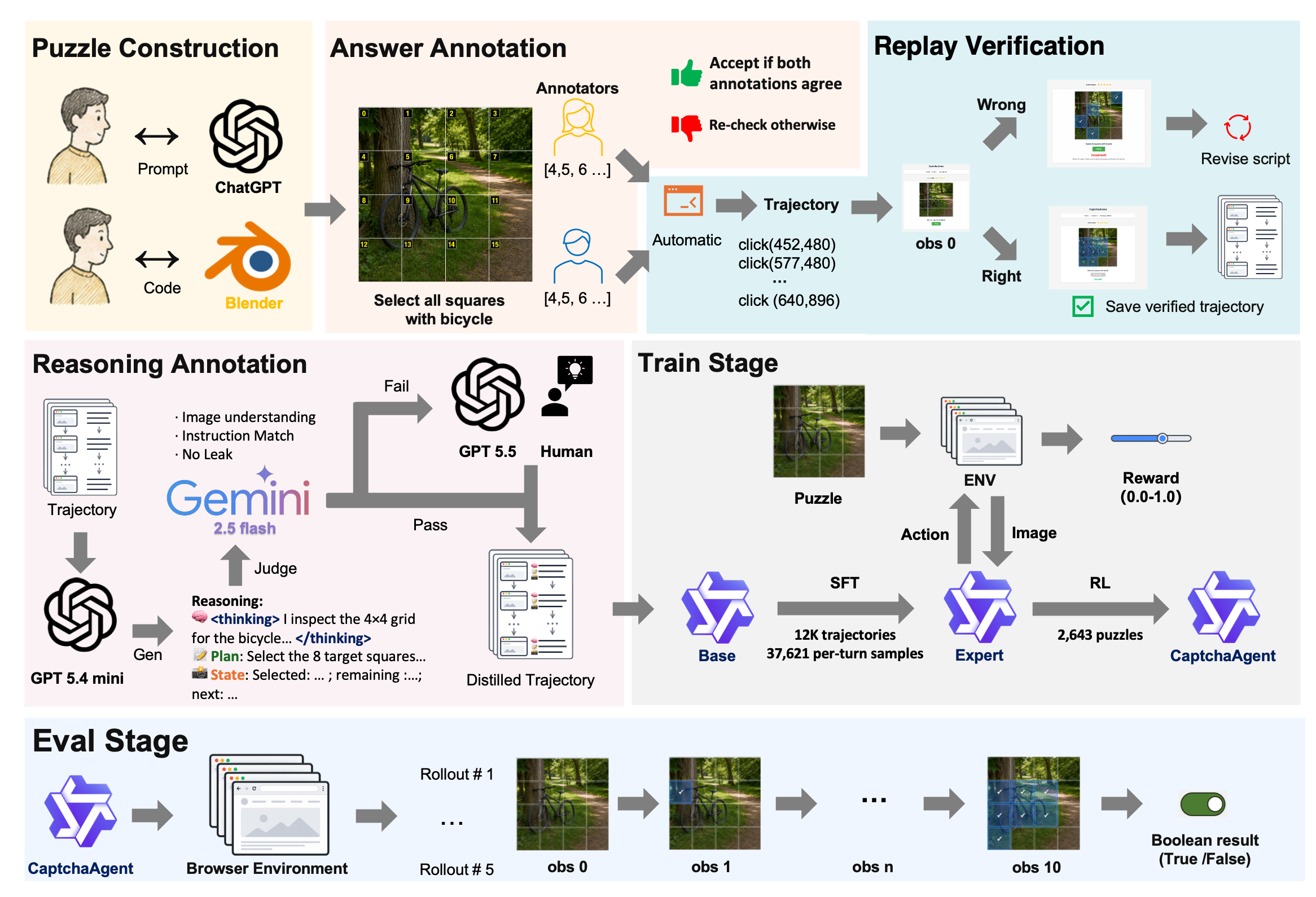}
\caption{Pipeline from puzzle construction and verification to reasoning annotation, training, and evaluation.}
\label{fig:pipeline}
\end{figure}

\textbf{CaptchaArena} is a large-scale, fine-grained computer-use training dataset for CAPTCHA solving in a live interactive environment. It contains $50$K puzzles across $20$ CAPTCHA types and $5$ interaction modes. Each type contributes $2{,}100$ training puzzles and $200$ each for validation and test, giving $42$K, $4$K, and $4$K puzzles, respectively (Table~\ref{tab:splits}). Unlike static CAPTCHA datasets that pair each challenge with an image and an answer, \textbf{CaptchaArena} provides closed-loop interaction supervision for training computer-use agents. Agents observe the page through a fixed $1280\times1080$ viewport and interact through low-level pointer and keyboard actions. Fig.~\ref{fig:pipeline} shows the pipeline from puzzle construction and verification to annotation, training, and evaluation. Every task requires visual recognition. Some tasks additionally require grounding for pixel-precise localization, logical reasoning, or mathematical computation (Table~\ref{tab:tasks}). Two authors annotated these skills independently, and the first author adjudicated disagreements.

\subsection{Puzzle Construction}

\paragraph{Action space.}
The agent has $5$ tool calls: screenshot, click$(x,y)$,
type\_text$(s)$, drag$(x_0,y_0,x_1,y_1)$ and timed hold$(x,y,\Delta t)$
(App.~\ref{app:action}). At each turn, it observes the current viewport with
the task prompt, emits one action, and the environment returns the next
screenshot. Drag and hold support interactions that cannot be expressed through clicks and text input alone. An episode ends when the puzzle is submitted. The agent presses submit, except in
$4$ types where submission occurs automatically after the solving action or when a timer expires. The same schema is used during replay,
annotation, supervised fine-tuning, reinforcement learning and evaluation.

\paragraph{Puzzle generation.}
Puzzles are generated under task-specific constraints.
Geometry Click and Path Finder are procedurally generated
using Blender,\footnote{\url{https://www.blender.org/}} which provides precise
control over geometry, viewpoint and lighting together with exact ground
truth. The images for most other types are generated with ChatGPT Images $2.0$ \citep{openai2026images2}. Three types use openly licensed icons and a public reCAPTCHA-V2 image set for part of their visual content (App.~\ref{app:generation}).
For images generated by ChatGPT, prompts diversify object identity, layout, visual style and distractors while preserving task semantics. Each task is described in App.~\ref{app:catalogue}. No proprietary CAPTCHA assets are scraped or redistributed.

\paragraph{Checking image quality.}
A generator does not always return the prompt request.
Dice Count shows several dice in one image, and the task is to
enter their total. For example, we may ask the model to generate an image whose dice sum to $23$, and
the dice in the returned image may sum to a different value. Every generated image is therefore inspected for
quality by two authors independently. An image is discarded and regenerated if either author rejects it. The two arithmetic types are checked further. Every Dice Count and Dart Count image is manually checked and labeled with the correct result. The rendered puzzle is checked separately
(App.~\ref{app:generation}).

\paragraph{Difficulty design.}
Where applicable, incorrect candidates are drawn from the same source as the correct candidate. They share its scene, style and framing, and differ only in the property being asked about. Answer positions are engineered as well. In procedurally generated types, positions are allocated by explicit quotas. Single-choice positions and arrow-cycle steps are near-uniform. The number of correct items in a multi-select task is uniform. Guessing a fixed count therefore offers no systematic advantage, and neither does guessing a fixed cycle position. Localization types are exempt, since their answers follow from the target itself (App.~\ref{app:quota}).

\begin{table}[t]
\centering
\footnotesize
\setlength{\tabcolsep}{6pt}
\renewcommand{\arraystretch}{1.05}
\caption{Dataset statistics for each CAPTCHA type.}
\label{tab:splits}
\begin{tabular}{@{}lrrr@{\hspace{18pt}}lrrr@{}}
\toprule
\textbf{Task Type} & \textbf{Train} & \textbf{Val} & \textbf{Test} &
\textbf{Task Type} & \textbf{Train} & \textbf{Val} & \textbf{Test} \\
\midrule
Geometry Click    & $2{,}100$ & $200$ & $200$ & Connect Icon   & $2{,}100$ & $200$ & $200$ \\
Misleading Click  & $2{,}100$ & $200$ & $200$ & Coordinates    & $2{,}100$ & $200$ & $200$ \\
Pick Area         & $2{,}100$ & $200$ & $200$ & Dart Count     & $2{,}100$ & $200$ & $200$ \\
Place Dot         & $2{,}100$ & $200$ & $200$ & Image Matching & $2{,}100$ & $200$ & $200$ \\
Select Animal     & $2{,}100$ & $200$ & $200$ & Object Match   & $2{,}100$ & $200$ & $200$ \\
Bingo             & $2{,}100$ & $200$ & $200$ & Path Finder    & $2{,}100$ & $200$ & $200$ \\
Click Order       & $2{,}100$ & $200$ & $200$ & Rotation Match & $2{,}100$ & $200$ & $200$ \\
Image Recognition & $2{,}100$ & $200$ & $200$ & Hold Button    & $2{,}100$ & $200$ & $200$ \\
Patch Select      & $2{,}100$ & $200$ & $200$ & Slide Puzzle   & $2{,}100$ & $200$ & $200$ \\
Unusual Detection & $2{,}100$ & $200$ & $200$ & Dice Count     & $2{,}100$ & $200$ & $200$ \\
\midrule
& & & & \textbf{Total} & $\mathbf{42{,}000}$ & $\mathbf{4{,}000}$ &
  $\mathbf{4{,}000}$ \\
\bottomrule
\end{tabular}
\end{table}

\subsection{Answer Annotation}
Ground truth comes from three sources: generation-time labels, manual annotation, and renderer-derived masks. Most types bind the image to its label at generation time, therefore the reference answer is computed by code rather than read from the rendered result. Types whose answers are a precise click or a set of selected patches are annotated by hand instead. Place Dot and Patch Select are independently annotated by two annotators, with the first author resolving disagreements. Localization types with irregular targets, such as Geometry Click, store a pixel mask. Irregular targets are evaluated against a mask rather than a bounding box (App.~\ref{app:egt}). For irregular-target tasks other than Misleading Click, the server accepts a click iff the corresponding mask pixel is set. Misleading Click uses the complementary rule, therefore its mask marks the region the agent must avoid. All $7{,}500$ puzzles of the $3$ irregular-target types carry a mask. Types whose target is a single point are exempt and use a stored coordinate with an explicit tolerance.

\subsection{Replay Verification}
Executing a puzzle's solution both verifies it and turns it into training data. A stored answer may be a semantic label, a coordinate, a set of coordinates, or a value. We compile it into an executable sequence of browser actions and run that sequence in the environment, recording a screenshot before every action. Every puzzle therefore yields one
screenshot-action trajectory. The same run also checks the answer. A puzzle is retained only if the final state is accepted by the page's verifier. A puzzle often admits more than one solution, and the same solution can be ordered in more than one way. Therefore, the environment can produce many trajectories for one puzzle, including ones that take a detour and recover from it. We release the environment and one accepted trajectory per puzzle.

\section{CaptchaAgent}
\label{sec:sft}

\textbf{CaptchaAgent} is a single policy over all $20$ types, trained in
two stages. The first stage is supervised fine-tuning on reasoning-annotated,
replay-verified trajectories from \S\ref{sec:data}. The
second stage is reinforcement learning with the environment verifier as reward. A single set of model parameters, tool schema, and observation format is shared across all task types, with no per-task adapter or task-conditioned switching. \S\ref{sec:distill} describes reasoning annotation,
\S\ref{sec:training} supervised fine-tuning, and \S\ref{sec:rl} reinforcement learning.

\subsection{Reasoning Annotation}
\label{sec:distill}

Replay yields one screenshot-action trajectory per puzzle (\S\ref{sec:data}). The actions are correct, but provide no explanation for why they were chosen. We therefore annotate every step with reasoning.

\paragraph{Teacher model.}
We give a single teacher (GPT-5.4-mini, \citealp{openai2026gpt54}) the screenshot at each turn with the correct action, and ask it to produce the reasoning that leads to that action. The action is described in task-level terms rather than as a coordinate, thus the teacher must locate the target in the image.

\paragraph{Judge filtering.}
An independent judge (Gemini-2.5-Flash, \citealp{comanici2025gemini25})  evaluates each generation twice. A
sample is admitted only if both checks accept it. The judge comes from a
different model family, so no model evaluates its own output. It inspects image
and text together and rejects on three criteria. \emph{Consistency.} Claims in the annotation must agree with the verified solution. \emph{No hindsight.} Answer-revealing phrasing is rejected, while first-person perceptual descriptions are allowed. \emph{Decisiveness.} Hedging and tentative language are rejected because the policy must commit to an action. Descriptions of icon and tile appearance are allowed because visual grounding is part of the annotation.

\paragraph{Additional verification.}
Rejected samples are returned to the teacher with the judge's feedback for
regeneration. Because the judge can also make visual errors, samples still
rejected after three attempts are checked by a stronger vision model (GPT-5.5, \citealp{openai2026gpt55}).  It also
checks every sample from Click Order and Connect Icon. In Click Order, weaker models may misread the glyphs, while in Connect Icon, they may miss the faint dashed links after an icon is moved. They may then misidentify which icons are connected. Samples disputed by the stronger model are regenerated under the same feedback loop. An author manually reviews, repairs, or discards any remaining cases.

\paragraph{Annotated data.}
Annotation covers the $42$K training and $4$K validation puzzles, giving
$46$K reasoning-annotated trajectories. Each trajectory carries the screenshot before every action, the action, and the
reasoning that leads to it. Because the chat template removes reasoning from previous turns, we expand
each $T$-turn trajectory into $T$ per-turn samples. Each turn is therefore supervised in the same format used at inference. Test puzzles are used only for evaluation and are never annotated.

\subsection{Supervised Fine-Tuning}
\label{sec:training}

We fine-tune on a multi-task subset spanning all $20$ types. We use $600$ of
each type's $2{,}100$ training puzzles, $12{,}000$ in total. Per-turn
expansion in \S\ref{sec:distill} gives $37{,}621$ training samples. The
remaining $1{,}500$ per type are held out and reserved for reinforcement learning on the
same environment. For SFT, the $200$ validation puzzles per type are used to monitor training, but
not for checkpoint selection.

The backbone is Qwen3.5-9B \citep{qwen35} with rank-$64$ LoRA adapters
\citep{hu2022lora} applied only to the language model. The vision encoder and the vision--language aligner are frozen. Training and evaluation use the same screenshot resolution and preprocessing. We train for $3$ epochs
and report the final checkpoint. App.~\ref{app:sftcfg} gives the full configuration.

\subsection{Reinforcement Learning}
\label{sec:rl}
Supervised fine-tuning leaves two failure modes unresolved. The policy does not always follow
a task through to a submission, and its actions do not always land precisely
enough for the verifier to accept them. We therefore continue training the
supervised policy with GRPO \citep{shao2024deepseekmath} against the live
environment. The environment verifier provides the reward. No reward model is learned, no preference data is collected, and no step-level reward annotation is required. The same verifier used to accept dataset solutions scores each rollout. Episodes are capped at $15$ turns.

Two design choices are important. First, difficulty is mined per puzzle rather than per type, since GRPO provides no learning signal when every rollout for a puzzle receives the same reward. Second, the reward is monotone both in submitting and in closeness to the answer, thus the policy is never rewarded for withholding an uncertain answer. App.~\ref{app:rl} gives the task pool, the reward formula, the optimization configuration, and the shortcuts observed before each reward term was introduced.

\section{Experiments}
\label{sec:results}

\subsection{Evaluation Setup}
\label{sec:setup}
We evaluate every system through the same interface. Each receives screenshots as input and predicts pixel-level actions, without an agent
framework, set-of-mark overlay or accessibility tree. The test set contains all $4{,}000$ puzzles across $20$ types, with $200$ puzzles per type. \textbf{CaptchaAgent} is sampled $5$ times per puzzle, and every other system
once. An episode ends when the page submits or the $15$-step cap is reached. It counts as solved only when the server-side verifier accepts the
submitted state. We report Pass@$k$ using the unbiased
estimator of \citet{chen2021codex}, including Pass@5 for \textbf{CaptchaAgent} (App.~\ref{app:passk}). Because every type contributes
the same number of test puzzles, the overall score is the unweighted mean
across types. We also report the full per-type breakdown.

\paragraph{Comparisons.}
We compare against the untrained Qwen3.5 backbone at $9$B, $27$B, and
$35$B-A$3$B, six open-weight GUI agents from $7$B to $9$B, three closed-source models, and a human reference of two annotators (App.~\ref{app:human}). The $9$B backbone differs from \textbf{CaptchaAgent} only in its weights.

\subsection{Results}
Table~\ref{tab:main} reports every type. The untrained $9$B backbone averages $11.4$ Pass@1, and scaling within the Qwen3.5 family to $27$B and $35$B-A$3$B does not improve performance (Table~\ref{tab:openbaselines}). The composition of this score is more informative than the average itself. The $9$B backbone averages $56.6$ on the $4$ types that submit automatically, but at most $0.5$ on the $16$ that require an explicit submit action. Its overall submit rate is $20.5\%$ (Table~\ref{tab:effort-open}). Its dominant failure is procedural rather than perceptual (App.~\ref{app:cases:modes}).

Across the evaluated systems, scale alone does not explain performance. The six open-weight GUI agents top out at $35.2$ (Table~\ref{tab:openbaselines}), while the three closed-source models, GPT 5.4 \citep{openai2026gpt54}, Gemini 3.5 Flash \citep{googledeepmind2026gemini35flash} and Claude Sonnet 4 \citep{anthropic2025claude4}, span $49.4$ to $69.2$. \textbf{CaptchaAgent} reaches $70.5$ average Pass@1 and $86.0$ Pass@5, with a $96.8\%$ submit rate. $8$ types reach $90$ or higher at Pass@1, $7$ of those sit within $5$ points of the human reference, and $4$ types in all match or exceed it. The remaining gap to the human average of $94.1$ is concentrated rather than uniform. Four types account for three-fifths of the $23.6$-point gap.

\begin{table}[t]
\centering
\scriptsize
\setlength{\tabcolsep}{3.5pt}
\caption{Per-type Pass@1 on the full test split, with $200$ puzzles per type.
Bold marks the best model result in each row; the human reference is excluded
from this comparison.}
\label{tab:main}
\begin{tabular}{@{}lrrrrrrrrr@{}}
\toprule
&\multicolumn{1}{c}{Untrained Base}  & \multicolumn{2}{c}{GUI agents} & \multicolumn{3}{c}{Closed-source}
& \multicolumn{2}{c}{\textbf{CaptchaAgent}} & \\
\cmidrule(lr){2-2}\cmidrule(lr){3-4}\cmidrule(lr){5-7}\cmidrule(lr){8-9}
Type & Qwen3.5-9B & Venus2-9B & GUI-Owl-8B & GPT-5.4 & Gemini & Claude & SFT & RL & Human \\
\midrule
Geometry Click     & 0.0  & \textbf{93.5} & 89.5 & 92.0 & 79.5 & 89.5 & 68.2 & 68.2 & 96.5 \\
Misleading Click   & 93.5 & 25.5 & 82.0 & 23.0 & 87.0 & 83.0 & \textbf{99.1} & 98.9 & 100.0 \\
Pick Area          & 35.5 & 23.0 & 33.0 & 37.5 & 37.5 & 36.0 & 59.8 & \textbf{60.6} & 97.5 \\
Place Dot          & 0.0  & 9.5  & 22.5 & \textbf{88.5} & 63.0 & 15.5 & 16.1 & 20.1 & 100.0 \\
Select Animal      & 0.0  & 28.5 & 99.5 & 98.5 & 90.5 & \textbf{100.0} & 99.3 & 99.4 & 98.0 \\
\midrule
Bingo              & 0.0  & 0.5  & 4.5  & 20.0 & \textbf{93.5} & 48.0 & 56.8 & 56.3 & 100.0 \\
Click Order        & 0.0  & 0.0  & 62.5 & 81.5 & 62.0 & 58.0 & \textbf{93.2} & 90.1 & 97.5 \\
Image Recognition  & 0.0  & 10.5 & 54.0 & \textbf{64.0} & 62.5 & 39.5 & 56.8 & 60.3 & 65.5 \\
Patch Select       & 0.0  & 1.0  & 5.0  & 9.0  & \textbf{12.5} & 12.0 & 9.6 & 10.4 & 73.5 \\
Unusual Detection  & 0.0  & 2.0  & 5.5  & 16.5 & 73.5 & 13.5 & 85.1 & \textbf{85.7} & 83.5 \\
\midrule
Connect Icon       & 0.0  & 48.5 & 13.5 & 45.0 & \textbf{90.5} & 60.5 & 71.7 & 76.8 & 98.5 \\
Coordinates        & 0.0  & 62.0 & 29.5 & 28.5 & 77.5 & 87.5 & 96.6 & \textbf{99.9} & 98.5 \\
Dart Count         & 0.0  & 99.0 & 28.0 & 43.5 & 82.5 & 94.5 & 99.0 & \textbf{99.2} & 98.0 \\
Image Matching     & 0.0  & 50.0 & 95.0 & 93.5 & 90.0 & 93.5 & 98.2 & \textbf{98.5} & 99.0 \\
Object Match       & 0.5  & 73.5 & 28.0 & 47.5 & 71.0 & 66.5 & 89.1 & \textbf{91.5} & 95.5 \\
Path Finder        & 0.0  & 66.5 & 14.0 & 54.5 & \textbf{78.0} & 49.5 & 71.9 & 72.9 & 100.0 \\
Rotation Match     & 0.0  & 3.5  & 10.0 & 22.0 & 70.0 & 40.5 & 91.5 & \textbf{94.5} & 100.0 \\
\midrule
Hold Button        & 97.5 & \textbf{100.0} & 0.0 & \textbf{100.0} & \textbf{100.0} & \textbf{100.0} & 99.9 & \textbf{100.0} & 98.0 \\
Slide Puzzle       & 0.0  & 1.0  & 2.0  & 13.5 & 40.0 & 16.5 & 42.2 & \textbf{43.9} & 99.5 \\
\midrule
Dice Count         & 0.0  & 6.5  & 1.0  & 9.0  & \textbf{22.5} & 7.5 & 5.7 & 7.3 & 83.0 \\
\midrule
\textbf{Average} & 11.4 & 35.2 & 34.0 & 49.4 & 69.2 & 55.6 & 70.5 &
  \textbf{71.7} & 94.1 \\
\bottomrule
\end{tabular}
\end{table}

\paragraph{Reinforcement learning.}
The best RL checkpoint improves average Pass@1 by $+1.2$ over the supervised policy,
or $+1.5$ excluding Click Order, with $16$ of $20$ types improving
(full paired Pass@$k$ in Table~\ref{tab:passk}). The
largest gains span the difficulty range rather than sitting at its bottom. Connect Icon improves by $+5.1$ from $71.7$, Place Dot by $+4.0$ from $16.1$, Image Recognition by $+3.5$ from $56.8$, and Coordinates by $+3.3$ from $96.6$, while the two weakest types move only slightly: Dice Count by $+1.6$ and Patch Select by $+0.8$. The improvement is larger at Pass@1 ($+1.2$) than at Pass@5 ($+0.8$). This pattern is consistent with RL improving first-rollout reliability more than expanding the set of puzzles the policy can solve across repeated attempts. Click Order is the only substantial regression, decreasing by $3.1$ points. RL also raises Pass@1 from $47.2$ to $51.0$ on Open CaptchaWorld and from $13.6$ to $20.0$ on Halligan (App.~\ref{app:ocw}).

\subsection{Error Analysis}

We diagnose failures with $3$ signals: submit rate, Pass@1, and Pass@5. A low submit rate shows that the policy often stops without finishing the episode. A low Pass@5 indicates that the policy cannot find a valid solution even across repeated attempts. A large gap between Pass@1 and Pass@5 indicates that a valid solution is reachable but not reliably reached on the first rollout. After SFT, $4$ types remain below $45$ Pass@1: Place Dot, Patch Select, Slide Puzzle, and Dice Count. All $4$ use all-or-nothing grading, where a single local error invalidates the episode. Their failures fall into $3$ patterns.

\paragraph{Non-submission.}
A rollout that never submits is scored as a failure regardless of page state. Untrained models submit on $20.0\%$ to $20.5\%$ of rollouts. Training largely removes the failure. The policy submits on $96.8\%$ of rollouts, and on at least $99\%$ in $13$ of the $20$ types (App.~\ref{app:effort}). Dice Count remains an exception. The supervised policy achieves only a $54.4\%$ submit rate, so nearly half its episodes are scored as failures with the answer never checked. The task itself is also hard. Humans reach $83.0$ here, their third-lowest score. The best system we evaluate reaches $22.5$. The $5.7$ Pass@1 therefore reflects both task difficulty and non-submission rather than pure counting ability.

\paragraph{Weak grounding.}
Two types remain low even across repeated attempts. Patch Select is
difficult for every evaluated system. No model exceeds $12.5$, while humans reach $73.5$. Place Dot is instead a specific weakness of \textbf{CaptchaAgent}. The policy submits on $95.3\%$ of rollouts but reaches only $16.1$ Pass@1, while GPT-5.4 reaches $88.5$ on the same puzzles. Five attempts
lift both types, to $27.5$ and $43.2$, but both remain far below their human references. Place Dot therefore exposes a remaining weakness in precise grounding.

\paragraph{Unstable execution.}
Some types reach valid solutions across repeated attempts but do so unreliably on the first rollout. The policy rises from $70.5$ at Pass@1 to $86.0$ at Pass@5 (App.~\ref{app:passk}), a $15.5$-point gap. The gap is not uniform. It exceeds $30$ points in Slide Puzzle ($42.2$ to $91.0$), Bingo ($56.8$ to $92.5$) and Pick Area ($59.8$ to $93.0$). However, it stays under $10$ points in the $8$ types already above $90$. These tasks therefore expose execution variance rather than uniformly missing capability.

\section{Conclusion}

We introduced \textbf{CaptchaArena}, a large-scale, fine-grained computer-use training dataset containing $50$K CAPTCHA puzzles across $20$ types.
Unlike static image--answer datasets, \textbf{CaptchaArena} represents
each puzzle as an executable browser task.
Every answer is a browser program, admitted only after its replay is accepted by the site's own verifier. The same process produces $50$K
screenshot-action trajectories, including $46$K with judge-filtered reasoning annotations. For irregular targets, \textbf{CaptchaArena} additionally provides pixel-mask supervision, allowing clicks to be graded against the target shape rather than a bounding box.

Using \textbf{CaptchaArena}, we further train \textbf{CaptchaAgent}, a single $9$B policy across all $20$ types. It reaches
$70.5$ Pass@1 after supervised fine-tuning and $71.7$ after reinforcement
learning using the same environment verifier as the reward. This compares with $11.4$ for the untrained
backbone, $35.2$ for the strongest open-weight GUI agent we evaluate, $69.2$ for the strongest closed-source model and $94.1$ for humans. Scaling the Qwen3.5 backbone does not close the gap, since a $35$B-A$3$B model also scores $11.4$, while the closed-source models span $49.4$ to $69.2$. These results point to training data and interaction supervision, rather than scale alone, as a key factor in CAPTCHA performance. The remaining gap to humans is concentrated in a small number of types and reflects three main failure modes: non-submission, weak grounding, and unstable execution. Because the environment deterministically verifies the
final page state, reinforcement learning requires neither a learned reward
model nor human reward labels.

\paragraph{Limitations.}
Reasoning annotations depend on commercial models, although they explain only replay-verified actions. The vision encoder stays frozen, and our largest gap to the human reference shows on a task requiring precise grounding. Actions are discrete, so trajectories cannot reproduce mouse-trajectory signals used by behavioral detectors \citep{acien2022becaptcha}.

\subsubsection*{Ethics statement}
\textbf{CaptchaArena} runs entirely in an environment we host ourselves. The external suites in App.~\ref{app:ocw} are also served locally. Halligan contains puzzles derived from CAPTCHA types deployed on real websites, but no
experiment interacts with a live CAPTCHA service or deployed anti-bot system,
and we neither collect nor replay traffic from one. \textbf{CaptchaArena} imagery is generated or openly licensed, with no proprietary assets redistributed (App.~\ref{app:generation}). The work studies CAPTCHA solving as a computer-use capability rather than bypassing protections on services we do not own. Because this capability is dual use, we release the dataset to
support both agent evaluation and the study of CAPTCHA robustness.

\subsubsection*{Reproducibility statement}
Each released puzzle includes an executable browser-action solution that can be replayed against the environment's own verifier. A third party can
therefore confirm that a released solution solves its puzzle without access to our generation pipeline (\S\ref{sec:data}). We do not claim that rerunning the stochastic generation process reconstructs the dataset exactly. \S\ref{sec:setup} specifies the evaluation protocol and Pass@$k$ estimator. The environment, the dataset and the model are released at \url{https://github.com/X0X0X00/CaptchaArena}.

\bibliographystyle{plainnat}
\bibliography{references}

\clearpage
\appendix
\section*{Appendix}
\section{Action Space and Coordinate Frames}
\label{app:action}

\paragraph{Action protocol.}
All agents use the same five-tool interface provided by the evaluation client: screenshot, click$(x,y)$, type\_text$(s)$,
drag$(x_0,y_0,x_1,y_1)$, and hold$(x,y,\Delta t)$. An episode ends when the page is submitted or when the $15$-turn limit is reached. Submission is explicit for most tasks. Geometry Click, Misleading Click, and Pick Area submit automatically after the solving click, while Hold Button submits when its timer expires.

\paragraph{Hold timing.}
\label{app:hold}
Hold Button requires the agent to hold a button for
$\Delta t \sim \mathrm{Unif}\{2,\dots,6\}$ seconds. The page checks the elapsed hold time every $100$\,ms and submits automatically once the required duration is reached. Its reference solution therefore contains a single hold$(x,y,\Delta t)$ action and no explicit submit action.

\paragraph{Coordinate frames.}
Ground truth and masks are stored in natural-image coordinates, with the origin at the image's top-left. Executable clicks are mostly in absolute pixels of the fixed $1280\times1080$ viewport. A point in natural-image coordinates is mapped to the viewport using the rendered element's bounding box $(x_0,y_0,w,h)$ and natural size $(W_n,H_n)$ as
\[
c_{\text{view}}
=
\left(
x_0 + \frac{w}{W_n}c_{\text{nat},x},
\;
y_0 + \frac{h}{H_n}c_{\text{nat},y}
\right).
\]
This allows the same annotation to be used consistently across render scales. The coordinate frame is stored per record rather than fixed by task type. Pick Area and Place Dot store targets in natural-image coordinates but executable clicks in viewport coordinates. Geometry Click stores both in natural-image coordinates and converts them at run time. Tasks built around fixed HTML controls use viewport coordinates only. The execution and evaluation code therefore reads the coordinate frame from each record.

\section{Task Catalogue}
\label{app:catalogue}
 Table~\ref{tab:tasks} gives each task's interaction mode,  the skills it needs beyond visual recognition, the length of its reference trajectory and its grading predicate. The paragraphs below group the tasks into $4$ families by page layout rather than by interaction mode. App.~\ref{app:egt} gives the exact acceptance predicates of the localization tasks.

\begin{table}[t]
\centering
\footnotesize
\setlength{\tabcolsep}{5pt}
\caption{The $20$ released \textbf{CaptchaArena} tasks grouped by
interaction mode. \emph{Mode}: S single-click, M multi-click, A arrow-cycle,
R real-time, T text-entry. \emph{Skills}: G grounding, L logic, M math. \emph{Action steps} gives the mean number of agent actions in the reference trajectory, rounded to one star per action and capped at five.}
\label{tab:tasks}
\begin{tabular}{@{}ccccccc@{}}
\toprule
\# & Task & Mode & Interaction & Skills & \shortstack{Action\\steps} & Grading \\
\midrule
1  & Geometry Click    & S & single-click                    & G  & \ding{72} & mask hit \\
2  & Misleading Click  & S & single-click                    &    & \ding{72} & mask \emph{miss} \\
3  & Pick Area         & S & single-click                    &    & \ding{72} & mask hit \\
4  & Place Dot         & S & single-click                    & G  & \ding{72}\ding{72} & distance tol. \\
5  & Select Animal     & S & single-click, $2{\times}3$ grid  &    & \ding{72}\ding{72} & target cell \\
\midrule
6  & Bingo              & M & multi-click, $3{\times}3$ grid   & L  & \ding{72}\ding{72}\ding{72} & any valid swap \\
7  & Click Order       & M & multi-click, ordered            & GL & \ding{72}\ding{72}\ding{72}\ding{72}\ding{72} & click order \\
8  & Image Recognition & M & multi-click, $3{\times}3$ grid   &    & \ding{72}\ding{72}\ding{72}\ding{72} & $A = A^\star$ \\
9  & Patch Select      & M & multi-click, $N{\times}N$ grid   &    & \ding{72}\ding{72}\ding{72}\ding{72}\ding{72} & $A = A^\star$ \\
10 & Unusual Detection & M & multi-click, $2{\times}3$ grid   &   & \ding{72}\ding{72}\ding{72}\ding{72} & $A = A^\star$ \\
\midrule
11 & Connect Icon      & A & arrow-cycle, $N{=}6$            &    & \ding{72}\ding{72}\ding{72}\ding{72} & $\hat k = k^\star$ \\
12 & Coordinates        & A & arrow-cycle, $N{=}5$            & L  & \ding{72}\ding{72}\ding{72} & $\hat k = k^\star$ \\
13 & Dart Count        & A & arrow-cycle, $N{=}5$            & M  & \ding{72}\ding{72}\ding{72}\ding{72} & $\hat k = k^\star$ \\
14 & Image Matching    & A & arrow-cycle, $N{=}5$            &    & \ding{72}\ding{72}\ding{72} & $\hat k = k^\star$ \\
15 & Object Match      & A & arrow-cycle, $N{=}5$            &   & \ding{72}\ding{72}\ding{72} & $\hat k = k^\star$ \\
16 & Path Finder       & A & arrow-cycle, $N{=}5$            & L  & \ding{72}\ding{72}\ding{72}\ding{72} & $\hat k = k^\star$ \\
17 & Rotation Match    & A & arrow-cycle, $8$ angles         & L  & \ding{72}\ding{72}\ding{72}\ding{72} & $\hat k = k^\star$ \\
\midrule
18 & Hold Button       & R & real-time hold                  &    & \ding{72} & duration interval \\
19 & Slide Puzzle      & R & real-time slider                &    & \ding{72}\ding{72}\ding{72} & distance tol. \\
\midrule
20 & Dice Count        & T & text entry                      & M  & \ding{72}\ding{72} & integer equality \\
\bottomrule
\end{tabular}
\end{table}

\subsection{Family A: Arrow-Cycle Single Choice}
The seven arrow-cycle tasks show a reference on the left and one candidate on the right. The agent uses arrow controls to cycle through candidates and submits the matching one.

\textbf{Image Matching} shows a type exemplar and five candidate images. Exactly one candidate belongs to the same class as the exemplar. The candidates share a similar scene and visual style, so the agent must identify the matching semantic type.

\textbf{Object Match} shows a counting card, for example "$3$ anchors" and $5$ scenes containing different numbers of the target object. The agent must count the objects and submit the scene matching the requested count.

\textbf{Coordinates} shows a seating chart in which rows are labeled with letters and columns with numbers. The reference specifies an address such as c$11$, and the agent must find the candidate chart that places Jerry at that location.

\textbf{Connect Icon} shows a reference pair of icons connected by a dashed line. Each candidate contains the same set of icons but connects a different pair. The agent cycles through the candidates and submits the one whose connection matches the reference.

\textbf{Path Finder} shows a $2$D path schematic with a marked location and several rendered views of the same scene, each placing a marker at a different position. The agent must transfer the marked location from the schematic to the rendered scene and select the matching candidate.

\textbf{Rotation Match} shows a hand pointing in one of $8$ directions. The arrow control rotates the subject image by $45^\circ$ at each step. The agent submits when the subject orientation matches the reference direction.

\textbf{Dart Count} shows a target sum and $5$ candidate dartboards. Each dartboard contains numbered regions, and only regions hit by a dart contribute to the total. The agent must identify the hit regions, sum their values, and submit the board matching the target sum.

\subsection{Family B: Single-Point Localization}
The four Family~B tasks are solved through a single image click. The page shows a single image and no candidate list. In three tasks, the click immediately submits the puzzle; Place Dot instead requires an explicit submit action. The grading rule depends on the task. Pick Area and Geometry Click accept clicks inside a target pixel mask, while Misleading Click rejects clicks inside a forbidden-region mask. Place Dot stores a target point and grades by distance. Pixel masks are used for irregular regions that cannot be represented accurately by bounding boxes.

\textbf{Pick Area} shows a scene divided into regions by drawn curves. The task is to click anywhere inside the largest region. Because the target region has an irregular shape, grading uses a pixel mask rather than a bounding box.

\textbf{Geometry Click} shows a rendered $3$D scene containing multiple objects. The prompt identifies one target by color and identity, for example "click yellow four". The agent must localize that object and click it. Grading uses a pixel mask because many targets, such as digits or curved shapes, are poorly represented by bounding boxes.

\textbf{Misleading Click} shows a character labeled "DON'T CLICK ME". The task is to click anywhere except the character itself. Grading uses a forbidden-region mask so that only clicks on the character are rejected, while background inside its bounding box remains valid.

\textbf{Place Dot} shows a city map with a path drawn across it. A car is parked at one endpoint, and the task is to click the other endpoint. The stored target is a point, and grading is based on distance to that point.

\subsection{Family C: Grid Multi-Select}
The four Family~C tasks show a grid of tiles and a prompt  specifying what to select. The agent selects one or more tiles and then presses submit. Grading uses exact set equality over the selected cells, so selecting an extra tile or missing a required tile causes failure.

\textbf{Image Recognition} shows a $3\times3$ grid of photographs and names a class, for example "select crosswalks". The task is to select every tile showing that class.

\textbf{Patch Select} overlays an $N\times N$ grid, with $N \in \{4,5\}$, on a photograph and names an object in the image. The task is to select every cell occupied by that object. Because objects need not align with grid boundaries, the agent must localize them precisely and judge borderline cells.

\textbf{Select Animal} shows a $2\times3$ grid of everyday objects and names one target, for example "click the camera." The task is to select the corresponding tile. Each puzzle has exactly one correct cell.

\textbf{Unusual Detection} shows a $2\times3$ grid of animals and asks the agent to select the unusual ones. An unusual tile is a chimera, such as the head of one animal combined with the body of another, while the remaining tiles show ordinary animals.

\subsection{Family D: Structured Singletons}
The five Family~D tasks each use a different interaction pattern rather than selecting among candidates. Four end when the agent presses submit, while Hold Button ends when the page's own timer completes.

\textbf{Bingo} shows a $3\times3$ board of icons and asks the agent to swap two tiles to complete a line of three identical icons. A swap consists of two clicks, one for each tile. A puzzle may admit multiple valid swaps, and any valid one is accepted.

\textbf{Click Order} shows a reference sequence of glyphs and an image containing scattered copies of those glyphs together with distractors. The task is to click the target glyphs in the specified order. The scattered glyphs may be colored, warped, rotated, or outlined, requiring the agent to recognize their identities rather than match their appearance directly.

\textbf{Slide Puzzle} shows an image with a missing piece and a horizontal slider. Dragging the slider moves the piece across the image. The agent must align the piece with the hole and then press submit.

\textbf{Hold Button} shows a button that must be held until a progress indicator completes. The required duration is $\Delta t \sim \mathrm{Unif}\{2,\dots,6\}$ seconds, and the page submits automatically once the threshold is reached (App.~\ref{app:hold}).

\textbf{Dice Count} shows $K \in \{1,2,4,6\}$ images containing multiple dice and asks for the total number of pips across all images. The agent computes the sum, enters it into a text field, and submits.

\section{Generation, Compositing, and Quality Control}
\label{app:generation}
\label{app:quota}

\paragraph{Asset provenance.}
Most generated images are produced with ChatGPT Images~2.0
\citep{openai2026images2}.
Connect Icon and Click Order additionally use an
openly licensed SVG icon collection.\footnote{\url{https://www.svgrepo.com/}} We
release the composited puzzle image, not the underlying icon
collection.
Image Recognition uses a public Google reCAPTCHA~V2 image dataset
released under CC0~1.0.\footnote{\url{https://www.kaggle.com/datasets/mikhailma/test-dataset}}
Geometry Click and Path Finder use Blender renders. Each CAPTCHA type maintains its own asset pool, although types requiring similar visual content may draw from overlapping source images.

\paragraph{Hand-annotated answers.}
Four types require answers to be read from the finished image rather than taken directly from the generator: Dice Count, Dart Count, Patch Select, and Place Dot. These correspond to the pip total, dart sum, occupied grid cells, and path endpoint, respectively.

\paragraph{Rendered-puzzle quality control.}
After checking all generated images, we randomly sample $200$ rendered puzzles per CAPTCHA type for manual inspection of formatting and rendering quality. Two authors independently check that the instruction, interaction elements, and reference answer are rendered correctly. Invalid or duplicate samples are discarded and regenerated. Disagreements are resolved by the first author.

\paragraph{Answers known by construction.}
For procedurally constructed tasks, some answers are available directly from the rendering process. The irregular-target localization tasks obtain their pixel masks from the generated target layer or renderer, rather than manual tracing. Grid-based tasks place known source tiles into known cells, so their answer sets are also determined during construction. Exact grading rules are given in App.~\ref{app:egt}.

\paragraph{Fixed layout and screenshot capture.}
Page layouts are fixed within each CAPTCHA type rather than randomly
repositioning controls across puzzles. Before each screenshot, the capture routine moves the cursor to $(5,5)$, away from all controls, preventing hover effects from a previous action from appearing as part of the next observation.

\paragraph{Answer balancing.}
Where applicable, answer positions are balanced across puzzles so that a fixed positional guess provides no systematic advantage. Single-choice and arrow-cycle tasks use near-uniform answer positions, while the number of correct items in multi-select tasks is also balanced. Localization tasks are excluded because their answers are determined by the target location itself.

\section{Certificate Schema and Acceptance Regions}
\label{app:egt}

\paragraph{Released schema.}
Each puzzle carries a raw semantic label and an executable certificate over the actions click, drag, type\_text, and hold. Bingo and the arrow-cycle tasks store a nested list of action sequences, where each member represents a valid reference solution. Only Bingo contains multiple sequences because a puzzle may admit several valid swaps; the arrow-cycle tasks contain a single sequence. The remaining tasks store a flat action sequence. A parser therefore reads both the nesting structure and each tool call's action field.

\paragraph{Acceptance predicates.}
The four localization tasks use different grading rules. Three use a stored pixel mask. A click $c$ is mapped to integer natural-image coordinates and tested as
\begin{equation}
\mathrm{hit}(c,M) = \mathbf{1}\big[M(\lfloor c \rceil) > 127\big].
\end{equation}
No morphology is applied at grading time.

\textbf{Geometry Click} accepts iff $\mathrm{hit}(c,M_{\mathrm{geo}})=1$. The generator dilates the target with a $10$\,px elliptical kernel before storing the mask, so the tolerance is encoded in the mask rather than added by the grader. The reference click is chosen as the deepest interior point of the undilated target.

\textbf{Pick Area} accepts iff $\mathrm{hit}(c,M_{\mathrm{pick}})=1$. Its mask is the generated target region
without dilation.

\textbf{Misleading Click} uses the complementary rule. Its mask marks the forbidden character, and any click outside that region is accepted.

\textbf{Place Dot} has a point target rather than a region and therefore does not use a mask. A click is accepted iff
$\lVert c-c^\star\rVert_2 < 12.5$ in natural-image pixels.

\paragraph{Why masks over rectangles.}
A bounding box can include background outside an irregular target, especially for concave or elongated shapes. A fixed-radius region can instead be too strict for large targets and too permissive for small ones. Pixel masks follow the target footprint directly and therefore support irregular shapes without these errors. Place Dot retains a distance-based rule because its target is a single point rather than a region. App.~\ref{app:cases:rect} measures the over-acceptance introduced by bounding boxes on evaluated clicks.

\section{Full Pass@\texorpdfstring{$k$}{k} Breakdown}
\label{app:passk}

Tables~\ref{tab:main} and~\ref{tab:openbaselines} report Pass@1 for every system. Here we report the full Pass@1 through Pass@5 curves for
both of our policies on the same test split Table~\ref{tab:passk}. Values are computed using the unbiased estimator in \S\ref{sec:setup} from five rollouts per puzzle. Other systems are omitted because they are evaluated with only one rollout per puzzle, for which only Pass@1 is available.

The gap between the two ends of a row is the quantity of interest. Where Pass@1 is low but Pass@5 is high, the policy can find the right answer within five attempts but cannot do so reliably on the first attempt. This indicates a variance problem rather than a failure of comprehension. When both Pass@1 and Pass@5 are low, additional attempts provide little benefit, indicating that the main limitation is what the policy can reliably discriminate. \S\ref{sec:results} discusses these patterns by task type.

\begin{table}[t]
\centering
\scriptsize
\caption{Per-type Pass@$k$ breakdown for the SFT and RL policies on the test split.}
\label{tab:passk}

\begin{subtable}{0.48\textwidth}
\centering
\setlength{\tabcolsep}{3pt}
\begin{tabular}{@{}lrrrrr@{}}
\toprule
Type & @$1$ & @$2$ & @$3$ & @$4$ & @$5$ \\
\midrule
Geometry Click     &  68.2 &  82.9 &  87.9 &  90.4 &  92.0 \\
Misleading Click   &  99.1 & 100.0 & 100.0 & 100.0 & 100.0 \\
Pick Area          &  59.8 &  78.1 &  86.1 &  90.3 &  93.0 \\
Place Dot          &  16.1 &  26.8 &  34.1 &  39.3 &  43.2 \\
Select Animal      &  99.3 & 100.0 & 100.0 & 100.0 & 100.0 \\
\midrule
Bingo              &  56.8 &  77.8 &  86.4 &  90.5 &  92.5 \\
Click Order        &  93.2 &  98.4 &  99.2 &  99.6 & 100.0 \\
Image Recognition  &  56.8 &  69.0 &  74.3 &  77.2 &  79.0 \\
Patch Select       &   9.6 &  16.4 &  21.2 &  24.8 &  27.5 \\
Unusual Detection  &  85.1 &  92.7 &  94.8 &  95.8 &  96.5 \\
\midrule
Connect Icon       &  71.7 &  89.1 &  95.0 &  97.4 &  98.5 \\
Coordinates        &  96.6 &  99.1 &  99.4 &  99.5 &  99.5 \\
Dart Count         &  99.0 &  99.5 &  99.5 &  99.5 &  99.5 \\
Image Matching     &  98.2 &  98.9 &  99.0 &  99.0 &  99.0 \\
Object Match       &  89.1 &  93.3 &  95.0 &  95.7 &  96.0 \\
Path Finder        &  71.9 &  84.5 &  89.3 &  91.9 &  93.5 \\
Rotation Match     &  91.5 &  97.3 &  99.0 &  99.7 & 100.0 \\
\midrule
Hold Button        &  99.9 & 100.0 & 100.0 & 100.0 & 100.0 \\
Slide Puzzle       &  42.2 &  65.0 &  78.1 &  86.0 &  91.0 \\
\midrule
Dice Count         &   5.7 &  10.2 &  13.9 &  17.1 &  20.0 \\
\midrule
\textbf{Average} & \textbf{70.5} & \textbf{79.0} & \textbf{82.6} &
  \textbf{84.7} & \textbf{86.0} \\
\bottomrule
\end{tabular}
\subcaption{Final SFT checkpoint (\S\ref{sec:training})}
\end{subtable}
\hfill
\begin{subtable}{0.48\textwidth}
\centering
\setlength{\tabcolsep}{3pt}
\begin{tabular}{@{}lrrrrr@{}}
\toprule
Type & @$1$ & @$2$ & @$3$ & @$4$ & @$5$ \\
\midrule
Geometry Click     &  68.2 &  78.9 &  83.3 &  86.0 &  88.0 \\
Misleading Click   &  98.9 & 100.0 & 100.0 & 100.0 & 100.0 \\
Pick Area          &  60.6 &  78.0 &  86.2 &  90.7 &  93.5 \\
Place Dot          &  20.1 &  31.7 &  39.2 &  44.7 &  49.2 \\
Select Animal      &  99.4 & 100.0 & 100.0 & 100.0 & 100.0 \\
\midrule
Bingo              &  56.3 &  79.2 &  89.3 &  94.0 &  96.5 \\
Click Order        &  90.1 &  97.6 &  98.8 &  99.0 &  99.0 \\
Image Recognition  &  60.3 &  71.9 &  76.8 &  79.8 &  82.0 \\
Patch Select       &  10.4 &  17.5 &  23.0 &  27.4 &  31.0 \\
Unusual Detection  &  85.7 &  93.6 &  95.8 &  96.9 &  97.5 \\
\midrule
Connect Icon       &  76.8 &  90.7 &  95.0 &  96.9 &  98.0 \\
Coordinates        &  99.9 & 100.0 & 100.0 & 100.0 & 100.0 \\
Dart Count         &  99.2 &  99.5 &  99.5 &  99.5 &  99.5 \\
Image Matching     &  98.5 &  99.0 &  99.3 &  99.4 &  99.5 \\
Object Match       &  91.5 &  95.0 &  96.1 &  96.6 &  97.0 \\
Path Finder        &  72.9 &  85.8 &  90.8 &  93.2 &  94.5 \\
Rotation Match     &  94.5 &  98.6 &  99.7 & 100.0 & 100.0 \\
\midrule
Hold Button        & 100.0 & 100.0 & 100.0 & 100.0 & 100.0 \\
Slide Puzzle       &  43.9 &  66.2 &  78.0 &  84.2 &  87.4 \\
\midrule
Dice Count         &   7.3 &  12.5 &  16.5 &  19.7 &  22.5 \\
\midrule
\textbf{Average} & \textbf{71.7} & \textbf{79.8} & \textbf{83.4} &
  \textbf{85.4} & \textbf{86.8} \\
\bottomrule
\end{tabular}
\subcaption{Best GRPO checkpoint (\S\ref{sec:rl})}
\end{subtable}
\end{table}

\section{Submit Rates}
\label{app:effort}

Table~\ref{tab:main} reports accuracy. Table~\ref{tab:openbaselines} extends the comparison to the two larger untrained backbones and all six open-weight GUI agents, while Tables~\ref{tab:effort-open} and~\ref{tab:effort-closed} report submit rates.

The untrained columns make the failure concrete. Outside the four types that submit automatically, the untrained models reach a graded submission on $0$ to $3\%$ of episodes. They act on the page and then stop without ever
committing to an answer, so almost none of their attempts are graded.
Place Dot is a clear example. The $9$B model submits on only $0.5\%$ of episodes, compared with $95.3\%$ for the supervised policy on the same puzzles.

\begin{table}[t]
\centering
\scriptsize
\setlength{\tabcolsep}{4pt}
\caption{Per-type Pass@1 for open-weight systems on the full test split.
The systems are Qwen3.5 \citep{qwen35}, UI-TARS-1.5-7B \citep{qin2025uitars, bytedance2025uitars15},
GUI-Owl-1.5-8B \citep{xu2026mobileagent}, UI-Venus-2-9B \citep{cai2026uivenus2},
Holo-3.1-9B \citep{hai2026holo31}, EvoCUA-8B \citep{xue2026evocua} and
Fara1.5-9B \citep{awadallah2026fara15}.}
\label{tab:openbaselines}
\begin{tabular}{@{}lrrrrrrrrr@{}}
\toprule
& \multicolumn{3}{c}{Qwen3.5} & \multicolumn{6}{c}{Open-weight GUI agents} \\
\cmidrule(lr){2-4}\cmidrule(l){5-10}
Type & 9B & 27B & 35B & UI-TARS & GUI-Owl & Venus2 & Holo-3.1 & EvoCUA & Fara1.5 \\
\midrule
Geometry Click     & 0.0    & 0.0    & 1.0    & 85.5   & 89.5   & \textbf{93.5} & 78.5   & 92.5   & 90.0   \\
Misleading Click   & \textbf{93.5} & \textbf{93.5} & 91.5   & 74.0   & 82.0   & 25.5   & 58.5   & 39.5   & 66.5   \\
Pick Area          & 35.5   & \textbf{38.5} & 34.5   & 33.0   & 33.0   & 23.0   & 27.5   & 29.5   & 30.0   \\
Place Dot          & 0.0    & 0.0    & 0.0    & 8.5    & 22.5   & 9.5    & 5.0    & \textbf{23.0} & 16.5   \\
Select Animal      & 0.0    & 0.0    & 0.0    & 93.0   & \textbf{99.5} & 28.5   & 80.0   & \textbf{99.5} & \textbf{99.5} \\
\midrule
Bingo              & 0.0    & 0.0    & 0.0    & 3.5    & 4.5    & 0.5    & 4.5    & 9.5    & \textbf{14.0} \\
Click Order        & 0.0    & 0.0    & 0.0    & 1.5    & \textbf{62.5} & 0.0    & 27.5   & 43.5   & 42.5   \\
Image Recognition  & 0.0    & 0.0    & 0.0    & 15.5   & \textbf{54.0} & 10.5   & 38.0   & 35.5   & 34.0   \\
Patch Select       & 0.0    & 0.0    & 0.0    & 0.5    & \textbf{5.0} & 1.0    & 1.0    & 1.5    & 4.0    \\
Unusual Detection  & 0.0    & 0.0    & 0.0    & 2.5    & 5.5    & 2.0    & \textbf{15.0} & 6.5    & 7.5    \\
\midrule
Connect Icon       & 0.0    & 0.0    & 0.0    & 1.5    & 13.5   & \textbf{48.5} & 11.0   & 15.5   & 7.5    \\
Coordinates        & 0.0    & 0.0    & 0.0    & 5.5    & 29.5   & \textbf{62.0} & 30.0   & 27.5   & 18.5   \\
Dart Count         & 0.0    & 0.0    & 0.0    & 3.0    & 28.0   & \textbf{99.0} & 61.0   & 22.5   & 19.0   \\
Image Matching     & 0.0    & 0.0    & 0.0    & 10.0   & \textbf{95.0} & 50.0   & 62.0   & 91.5   & 63.0   \\
Object Match       & 0.5    & 0.0    & 0.0    & 15.5   & 28.0   & \textbf{73.5} & 28.0   & 34.0   & 29.5   \\
Path Finder        & 0.0    & 0.0    & 0.0    & 12.5   & 14.0   & \textbf{66.5} & 25.5   & 22.0   & 30.5   \\
Rotation Match     & 0.0    & 0.0    & 0.0    & 3.5    & 10.0   & 3.5    & 6.5    & 10.5   & \textbf{19.0} \\
\midrule
Hold Button        & 97.5   & \textbf{100.0} & \textbf{100.0} & 0.0    & 0.0    & \textbf{100.0} & 76.0   & 0.0    & 0.0    \\
Slide Puzzle       & 0.0    & 0.0    & 0.0    & \textbf{2.5} & 2.0    & 1.0    & 0.5    & 0.0    & 1.0    \\
\midrule
Dice Count         & 0.0    & 0.0    & 0.0    & 0.0    & 1.0    & \textbf{6.5} & 5.5    & 1.5    & 2.0    \\
\midrule
\textbf{Average} & 11.4 & 11.6 & 11.4 & 18.6 & 34.0 & \textbf{35.2} & 32.1 & 30.3 & 29.7 \\
\bottomrule
\end{tabular}

\end{table}

\begin{table}[t]
\centering
\scriptsize
\setlength{\tabcolsep}{4pt}
\caption{Submit rate (\%) for open-weight systems. A rollout that never submits is scored as a failure, so the submit rate places an upper bound on Pass@1. Pass@1 for the same systems is in Table~\ref{tab:openbaselines}.}
\label{tab:effort-open}
\begin{tabular}{@{}lrrrrrrrrr@{}}
\toprule
& \multicolumn{3}{c}{Qwen3.5} & \multicolumn{6}{c}{Open-weight GUI agents} \\
\cmidrule(lr){2-4}\cmidrule(l){5-10}
Type & 9B & 27B & 35B & UI-TARS & GUI-Owl & Venus2 & Holo-3.1 & EvoCUA & Fara1.5 \\
\midrule
Geometry Click     & 100.0  & 99.5   & 99.0   & 100.0  & 100.0  & 100.0  & 99.5   & 99.0   & 100.0  \\
Misleading Click   & 99.5   & 100.0  & 100.0  & 100.0  & 100.0  & 100.0  & 97.0   & 100.0  & 100.0  \\
Pick Area          & 100.0  & 100.0  & 100.0  & 99.5   & 100.0  & 100.0  & 94.0   & 100.0  & 100.0  \\
Place Dot          & 0.5    & 0.0    & 0.0    & 67.5   & 100.0  & 52.5   & 56.5   & 100.0  & 98.0   \\
Select Animal      & 0.0    & 0.0    & 0.0    & 95.0   & 99.5   & 29.0   & 80.0   & 100.0  & 100.0  \\
\midrule
Bingo              & 2.5    & 0.0    & 0.0    & 50.0   & 73.5   & 1.0    & 32.0   & 84.5   & 92.5   \\
Click Order        & 0.0    & 0.0    & 0.0    & 45.5   & 94.5   & 1.5    & 56.5   & 81.5   & 66.5   \\
Image Recognition  & 1.0    & 0.0    & 0.5    & 92.0   & 100.0  & 19.5   & 94.5   & 100.0  & 97.0   \\
Patch Select       & 0.5    & 0.0    & 0.0    & 59.5   & 76.5   & 6.5    & 46.5   & 96.0   & 92.0   \\
Unusual Detection  & 0.5    & 0.0    & 0.0    & 81.5   & 98.5   & 10.0   & 63.0   & 100.0  & 96.5   \\
\midrule
Connect Icon       & 3.0    & 0.0    & 0.5    & 17.5   & 78.0   & 97.0   & 63.5   & 77.5   & 32.5   \\
Coordinates        & 1.0    & 0.0    & 0.0    & 34.0   & 81.5   & 84.5   & 63.5   & 96.5   & 60.0   \\
Dart Count         & 1.0    & 0.0    & 0.0    & 19.5   & 69.5   & 100.0  & 73.0   & 52.5   & 66.0   \\
Image Matching     & 1.5    & 0.0    & 0.0    & 22.5   & 99.0   & 64.5   & 73.0   & 98.5   & 80.0   \\
Object Match       & 1.5    & 0.0    & 0.0    & 48.0   & 88.5   & 91.0   & 59.5   & 94.5   & 85.0   \\
Path Finder        & 0.0    & 0.0    & 0.0    & 58.5   & 98.0   & 94.0   & 70.0   & 92.0   & 97.0   \\
Rotation Match     & 0.0    & 0.0    & 0.0    & 59.0   & 90.5   & 29.0   & 35.0   & 91.0   & 91.5   \\
\midrule
Hold Button        & 97.5   & 100.0  & 100.0  & 0.0    & 0.0    & 100.0  & 76.0   & 0.0    & 0.0    \\
Slide Puzzle       & 0.0    & 0.0    & 0.0    & 14.0   & 82.5   & 11.0   & 62.5   & 19.5   & 34.5   \\
\midrule
Dice Count         & 0.0    & 0.0    & 0.0    & 48.5   & 33.5   & 51.5   & 72.5   & 51.0   & 92.5   \\
\midrule
\textbf{Average} & \textbf{20.5} & \textbf{20.0} & \textbf{20.0} & \textbf{55.6} & \textbf{83.2} & \textbf{57.1} & \textbf{68.4} & \textbf{81.7} & \textbf{79.1} \\
\bottomrule
\end{tabular}
\end{table}

\begin{table}[t]
\centering
\footnotesize
\setlength{\tabcolsep}{8pt}
\caption{Submit rate (\%) for the closed-source models and \textbf{CaptchaAgent}. Closed-source models use $1$ rollout per puzzle, while \emph{SFT} and \emph{RL} use $5$.}
\label{tab:effort-closed}
\begin{tabular}{@{}lrrrrr@{}}
\toprule
& \multicolumn{3}{c}{Closed-source models} & \multicolumn{2}{c}{\textbf{CaptchaAgent}} \\
\cmidrule(lr){2-4}\cmidrule(l){5-6}
Type & GPT-5.4 & Gemini & Claude & SFT & RL \\
\midrule
Geometry Click    & 100.0 & 100.0 & 100.0 & 99.9  & 99.9 \\
Misleading Click  & 100.0 & 99.0  & 100.0 & 99.7  & 99.6 \\
Pick Area         & 99.0  & 98.0  & 100.0 & 100.0 & 100.0 \\
Place Dot         & 100.0 & 67.5  & 100.0 & 95.3  & 99.8 \\
Select Animal     & 100.0 & 93.0  & 100.0 & 99.9  & 99.8 \\
\midrule
Bingo             & 100.0 & 94.0  & 99.5  & 100.0 & 100.0 \\
Click Order       & 100.0 & 71.5  & 95.0  & 96.9  & 97.3 \\
Image Recognition & 100.0 & 96.5  & 100.0 & 99.8  & 100.0 \\
Patch Select      & 100.0 & 65.0  & 100.0 & 98.5  & 89.9 \\
Unusual Detection & 100.0 & 88.0  & 100.0 & 99.9  & 99.8 \\
\midrule
Connect Icon      & 100.0 & 91.0  & 97.5  & 99.8  & 99.9 \\
Coordinates       & 100.0 & 78.0  & 91.0  & 97.4  & 100.0 \\
Dart Count        & 100.0 & 85.0  & 97.0  & 99.9  & 99.8 \\
Image Matching    & 100.0 & 90.5  & 100.0 & 99.2  & 99.3 \\
Object Match      & 100.0 & 74.0  & 96.5  & 98.0  & 98.2 \\
Path Finder       & 100.0 & 86.0  & 100.0 & 99.9  & 100.0 \\
Rotation Match    & 100.0 & 79.5  & 100.0 & 98.2  & 100.0 \\
\midrule
Hold Button       & 100.0 & 100.0 & 100.0 & 99.9  & 100.0 \\
Slide Puzzle      & 99.0  & 96.5  & 98.5  & 99.8  & 99.0 \\
\midrule
Dice Count        & 100.0 & 72.0  & 100.0 & 54.4  & 76.6 \\
\midrule
\textbf{Average} & \textbf{99.9} & \textbf{86.3} & \textbf{98.8} &
\textbf{96.8} & \textbf{97.9} \\
\bottomrule
\end{tabular}
\end{table}

\section{Human Reference}
\label{app:human}

Two annotators solved all $4{,}000$ test puzzles over $10$ days, each
completing $200$ randomly assigned puzzles per day, so every puzzle was solved exactly once. They used the same rendered page, fixed viewport, and interaction interface as the agents, with no answer key, hints, or second attempt. A puzzle counts as solved only when the page's verifier accepts the first submission. Neither annotator was familiar with the task formats beforehand. We report the median time per puzzle because occasional idle periods can substantially inflate the mean.

The gap to the supervised policy is concentrated in a few task types.
Place Dot ($-83.9$), Dice Count ($-77.3$), Patch Select ($-63.9$), and Slide Puzzle ($-57.3$) together account for $14.1$ points of the
$23.6$-point average gap. The policy matches or exceeds the annotators on Select Animal, Unusual Detection, Dart Count, and Hold Button. Reinforcement learning reduces the average gap from $23.6$ to $22.4$ points but leaves the same four largest deficits. A low human score has a different interpretation from a low policy score. Image Recognition has the lowest human accuracy ($65.5$), suggesting genuine ambiguity about which cells contain the target. In contrast, annotators solve Place Dot perfectly and Slide Puzzle in $199$ of $200$ cases, showing that the large policy deficits on these tasks cannot be explained by task ambiguity.

\begin{table}[t]
\centering
\footnotesize
\setlength{\tabcolsep}{5pt}
\caption{Human reference on $20$ types at
$200$ puzzles each, against both policies.}
\label{tab:human}
\begin{tabular}{@{}llrrrrrr@{}}
\toprule
& & \multicolumn{2}{c}{Human} & \multicolumn{2}{c}{Pass@1} &
\multicolumn{2}{c}{$\Delta$} \\
\cmidrule(lr){3-4}\cmidrule(lr){5-6}\cmidrule(l){7-8}
Mode & Type & Acc. & Time (s) & SFT & RL & SFT & RL \\
\midrule
\multirow{5}{*}{Single-click}
 & Geometry Click & 96.5 & 6.1 & 68.2 & 68.2 & $-28.3$ & $-28.3$ \\
 & Misleading Click & 100.0 & 3.3 & 99.1 & 98.9 & $-0.9$ & $-1.1$ \\
 & Pick Area & 97.5 & 2.4 & 59.8 & 60.6 & $-37.7$ & $-36.9$ \\
 & Place Dot & 100.0 & 4.8 & 16.1 & 20.1 & $-83.9$ & $-79.9$ \\
 & Select Animal & 98.0 & 5.7 & 99.3 & 99.4 & $+1.3$ & $+1.4$ \\
\midrule
\multirow{5}{*}{Multi-click}
 & Bingo & 100.0 & 4.9 & 56.8 & 56.3 & $-43.2$ & $-43.7$ \\
 & Click Order & 97.5 & 11.0 & 93.2 & 90.1 & $-4.3$ & $-7.4$ \\
 & Image Recognition & 65.5 & 7.4 & 56.8 & 60.3 & $-8.7$ & $-5.2$ \\
 & Patch Select & 73.5 & 5.3 & 9.6 & 10.4 & $-63.9$ & $-63.1$ \\
 & Unusual Detection & 83.5 & 7.5 & 85.1 & 85.7 & $+1.6$ & $+2.2$ \\
\midrule
\multirow{7}{*}{Arrow-cycle}
 & Connect Icon & 98.5 & 9.3 & 71.7 & 76.8 & $-26.8$ & $-21.7$ \\
 & Coordinates & 98.5 & 5.1 & 96.6 & 99.9 & $-1.9$ & $+1.4$ \\
 & Dart Count & 98.0 & 19.5 & 99.0 & 99.2 & $+1.0$ & $+1.2$ \\
 & Image Matching & 99.0 & 18.9 & 98.2 & 98.5 & $-0.8$ & $-0.5$ \\
 & Object Match & 95.5 & 5.0 & 89.1 & 91.5 & $-6.4$ & $-4.0$ \\
 & Path Finder & 100.0 & 4.9 & 71.9 & 72.9 & $-28.1$ & $-27.1$ \\
 & Rotation Match & 100.0 & 5.4 & 91.5 & 94.5 & $-8.5$ & $-5.5$ \\
\midrule
\multirow{2}{*}{Real-time}
 & Hold Button & 98.0 & 3.9 & 99.9 & 100.0 & $+1.9$ & $+2.0$ \\
 & Slide Puzzle & 99.5 & 9.2 & 42.2 & 43.9 & $-57.3$ & $-55.6$ \\
\midrule
Text entry
 & Dice Count & 83.0 & 30.8 & 5.7 & 7.3 & $-77.3$ & $-75.7$ \\
\midrule
\multicolumn{2}{@{}l}{\textbf{Average}}
  & \textbf{94.1} & \textbf{8.5} & \textbf{70.5} & \textbf{71.7} &
  $\mathbf{-23.6}$ & $\mathbf{-22.4}$ \\
\bottomrule
\end{tabular}
\\[3pt]
{\scriptsize
\quad \emph{Time}: the median seconds per puzzle. \quad
$\Delta$: a policy's Pass@1 minus the human accuracy, thus a positive value
means the policy is ahead.}
\end{table}

\section{Source-Asset Reuse Across Splits}
\label{app:reuse}
Where a type's asset pool is large enough, splits are constructed by
partitioning source assets before rendering. Types built from smaller pools cannot always be partitioned exactly, so we audited source-image overlap between training and test using content hashes rather than filenames. This distinction matters: in Unusual Detection, all $200$ test filenames also occur in training, while the content overlap is exactly zero. Under content hashing, $12$ of the $20$ types share no source image between training and test. Six types share source images: Image Matching ($276$ of $282$ test
images), Coordinates ($113$ of $192$), Dart Count ($101$ of $293$), Misleading Click ($95$ of $333$), Patch Select ($48$ of
$200$), and Rotation Match ($32$ of $232$). Image Recognition stores its cells in per-puzzle subdirectories not covered by the scan, so we report no overlap figure for it. We also omit Hold Button, since the reused asset is only the button image, while the task itself is
determined by the required hold duration.

Source-image reuse does not imply puzzle duplication. The target, distractors, layout, and answer are sampled per puzzle, so a reused image can appear in a different task instance with a different answer. We report these measurements so that readers can account for source reuse when interpreting generalization results.

\section{Generalization Analysis}
\label{app:ocw}
We evaluate on two benchmarks built independently of \textbf{CaptchaArena}. Open CaptchaWorld \citep{luo2025opencaptchaworld} contains $463$ puzzles across the same $20$ task types that we adopt, but uses its own renderers,
layouts, and source assets. Halligan \citep{teoh2025halligan} contains $2{,}600$ puzzles across $26$ CAPTCHA types derived from real websites. Some share similar high-level task formats with our benchmark, but differ in layout, visual content, and task instantiation. Neither benchmark contains data from our generation pipeline or validation split.

\paragraph{Protocol.}
We serve both benchmarks locally and evaluate them through the same
screenshot-based agent interface, $15$-turn cap, and $1280\times1080$
viewport used throughout this paper. Open CaptchaWorld was originally
evaluated through browser-use, so our conversion replaces structured
element-level actions with pixel coordinates that the policy must localize itself. Published scores are therefore not directly comparable with ours because the interaction and evaluation protocols differ. Every system is evaluated with one rollout per puzzle.

\begin{table}[t]
\centering
\footnotesize
\setlength{\tabcolsep}{6pt}
\caption{Generalization to two external evaluation benchmarks. Best results for each metric are shown in bold.}
\label{tab:ood}
\begin{tabular}{@{}lrrrrrr@{}}
\toprule
& \multicolumn{3}{c}{Open CaptchaWorld} & \multicolumn{3}{c}{Halligan} \\
\cmidrule(lr){2-4}\cmidrule(l){5-7}
Model & Pass@1 & Submit & Steps & Pass@1 & Submit & Steps \\
\midrule
Qwen3.5-9B & 9.8 & 21.1 & 9.5 & 0.9 & 12.6 & 12.7 \\
UI-TARS-1.5-7B & 14.0 & 53.8 & 9.1 & 0.9 & 13.2 & 12.7 \\
\midrule
SFT (\S\ref{sec:training}) & 47.2 & 90.6 & 4.2 & 13.6 & 53.4 & 8.7 \\
RL (\S\ref{sec:rl}) & \textbf{51.0} & \textbf{94.0} & \textbf{3.9} &
  \textbf{20.0} & \textbf{66.9} & \textbf{7.2} \\
\bottomrule
\end{tabular}
\\[3pt]
{\scriptsize
One rollout per puzzle.
\emph{Submit}: submission rate (\%).
\emph{Steps}: mean number of executed actions per puzzle.
Bold marks the highest Pass@1 in each benchmark.}
\end{table}

\paragraph{Results.}
Reinforcement learning improves performance on both external benchmarks, by $+3.8$ Pass@1 on Open CaptchaWorld and $+6.4$ on Halligan, compared with $+1.2$ on our own test split. Submit rates also increase while the mean number of executed actions decreases on both benchmarks (Table~\ref{tab:ood}).

On Open CaptchaWorld, RL improves on $9$ types, declines on $5$, and leaves $6$ unchanged (Table~\ref{tab:ocwcat}). The largest gains are Path Finder ($+20.0$) and Coordinates ($+16.7$), while the largest drops are Dart Count ($-10.0$) and Rotation Match ($-8.3$).

On Halligan, RL improves on $14$ of $26$ types, declines on $8$, and leaves $4$ unchanged (Table~\ref{tab:halcat}). About $70\%$ of the aggregate improvement comes from yandex/text ($+67.0$), botdetect ($+29.0$), and mtcaptcha ($+25.0$). Absolute transfer remains limited at $20.0$; none of Halligan's $26$ types appears in our training corpus.

Two UI-TARS results reflect interface limitations rather than CAPTCHA
recognition. On Halligan, it repeatedly clicks the same coordinate on the Arkose and FunCAPTCHA start screens and reaches the puzzle in only $11$ of $1{,}200$ episodes, producing $0.0$ on all $12$ corresponding types. On Open CaptchaWorld, it scores $0.0$ on Hold Button because its action space does not support a timed hold.

\begin{table}[t]
\centering
\scriptsize
\caption{Per-type Pass@1 on the two external evaluation benchmarks.}
\begin{subtable}[t]{0.47\textwidth}
\centering
\setlength{\tabcolsep}{2.5pt}
\begin{tabular}{@{}lrrrrr@{}}
\toprule
Type & $n$ & Qwen & TARS & SFT & RL \\
\midrule
Bingo             & $25$ & 0.0   & 0.0  & 44.0  & 40.0 \\
Click Order       & $20$ & 0.0   & 5.0  & 45.0  & 45.0 \\
Connect Icon      & $20$ & 0.0   & 0.0  & 60.0  & 60.0 \\
Coordinates       & $18$ & 0.0   & 0.0  & 44.4  & 61.1 \\
Dart Count        & $20$ & 0.0   & 5.0  & 75.0  & 65.0 \\
Dice Count        & $20$ & 0.0   & 10.0 & 0.0   & 15.0 \\
Geometry Click    & $20$ & 0.0   & 15.0 & 5.0   & 20.0 \\
Hold Button       & $10$ & 100.0 & 0.0  & 100.0 & 100.0 \\
Image Matching    & $19$ & 0.0   & 15.8 & 94.7  & 94.7 \\
Image Recognition & $20$ & 0.0   & 30.0 & 65.0  & 70.0 \\
Misleading Click  & $20$ & 75.0  & 30.0 & 90.0  & 85.0 \\
Object Match      & $20$ & 0.0   & 15.0 & 60.0  & 55.0 \\
Patch Select      & $20$ & 0.0   & 0.0  & 0.0   & 10.0 \\
Path Finder       & $10$ & 0.0   & 10.0 & 30.0  & 50.0 \\
Pick Area         & $30$ & 20.0  & 20.0 & 46.7  & 46.7 \\
Place Dot         & $32$ & 0.0   & 0.0  & 0.0   & 3.1 \\
Rotation Match    & $48$ & 0.0   & 16.7 & 52.1  & 43.8 \\
Select Animal     & $30$ & 0.0   & 93.3 & 86.7  & 100.0 \\
Slide Puzzle      & $31$ & 0.0   & 0.0  & 29.0  & 38.7 \\
Unusual Detection & $30$ & 0.0   & 13.3 & 16.7  & 16.7 \\
\midrule
\textbf{Average} & $--$ & \textbf{9.8} & \textbf{14.0} & \textbf{47.2} & \textbf{51.0} \\
\bottomrule
\end{tabular}
\subcaption{Open CaptchaWorld}
\label{tab:ocwcat}
\end{subtable}\hfill
\begin{subtable}[t]{0.5\textwidth}
\centering
\setlength{\tabcolsep}{2.5pt}
\begin{tabular}{@{}lrrrr@{}}
\toprule
Type & Qwen & TARS & SFT & RL \\
\midrule
amazon/waf              & 0.0 & 0.0  & 8.0  & 6.0 \\
arkose/3d\_rollball     & 0.0 & 0.0  & 2.0  & 2.0 \\
arkose/dice\_match      & 0.0 & 0.0  & 21.0 & 20.0 \\
arkose/numbermatch      & 0.0 & 0.0  & 2.0  & 6.0 \\
arkose/orbit\_match     & 0.0 & 0.0  & 23.0 & 21.0 \\
arkose/rockstack        & 0.0 & 0.0  & 3.0  & 4.0 \\
baidu/rotate            & 0.0 & 7.0  & 2.0  & 8.0 \\
botdetect               & 1.0 & 11.0 & 1.0  & 30.0 \\
funcaptcha/card         & 1.0 & 0.0  & 19.0 & 11.0 \\
funcaptcha/counting     & 3.0 & 0.0  & 13.0 & 17.0 \\
funcaptcha/dice\_pair   & 4.0 & 0.0  & 11.0 & 20.0 \\
funcaptcha/galaxies     & 0.0 & 0.0  & 74.0 & 71.0 \\
funcaptcha/hand\_number & 2.0 & 0.0  & 10.0 & 16.0 \\
funcaptcha/rotated      & 3.0 & 0.0  & 29.0 & 39.0 \\
funcaptcha/square\_icon & 3.0 & 0.0  & 35.0 & 42.0 \\
geetest/gobang          & 1.0 & 0.0  & 0.0  & 0.0 \\
geetest/icon            & 0.0 & 0.0  & 31.0 & 25.0 \\
geetest/iconcrush       & 4.0 & 1.0  & 9.0  & 16.0 \\
geetest/slide           & 0.0 & 2.0  & 4.0  & 19.0 \\
hcaptcha                & 0.0 & 0.0  & 6.0  & 5.0 \\
lemin                   & 0.0 & 0.0  & 1.0  & 0.0 \\
mtcaptcha               & 0.0 & 0.0  & 14.0 & 39.0 \\
recaptchav2             & 0.0 & 0.0  & 1.0  & 2.0 \\
tencent/vtt             & 0.0 & 1.0  & 30.0 & 30.0 \\
yandex/kaleidoscope     & 0.0 & 1.0  & 1.0  & 1.0 \\
yandex/text             & 1.0 & 0.0  & 3.0  & 70.0 \\
\midrule
\textbf{Average} & \textbf{0.9} & \textbf{0.9} & \textbf{13.6} & \textbf{20.0} \\
\bottomrule
\end{tabular}
\subcaption{Halligan}
\label{tab:halcat}
\end{subtable}
\\[3pt]
{\scriptsize
One rollout per puzzle.
\emph{Qwen}: untrained Qwen3.5-9B.
\emph{TARS}: UI-TARS-1.5-7B.
\emph{Average}: unweighted mean over task types.
Open CaptchaWorld contains $463$ puzzles with uneven type sizes;
Halligan contains $100$ puzzles per type.}
\end{table}

\section{Case Studies}
\label{app:cases}

Two claims in the main text are examined here on individual rollouts. First, grading a click against a bounding rectangle can accept clicks that miss an irregular target (\S\ref{sec:intro}). App.~\ref{app:egt} explains this geometrically, and App.~\ref{app:cases:rect} measures the effect on clicks made during evaluation. Second, \S\ref{sec:results} identifies procedural failures that are not explained by visual recognition alone. App.~\ref{app:cases:modes} illustrates these failures through examples from UI-TARS-1.5-7B and \textbf{CaptchaAgent}.

Every rollout below comes from the evaluation runs behind
Tables~\ref{tab:main} and~\ref{tab:openbaselines}. We use the same UI-TARS rollouts reported there.

\subsection{Rectangle Grading Inflates Click Accuracy}
\label{app:cases:rect}

\paragraph{Set-up.}
Geometry Click and Pick Area are the two types whose targets are irregular regions graded by a stored mask (App.~\ref{app:egt}). Misleading Click, the third irregular-target type in \S\ref{sec:data}, is excluded because its mask marks the region a click must avoid, so a bounding rectangle could only reject more clicks rather than accept more. For every evaluated click, we map the coordinate to natural-image pixels exactly as the front-end does and grade it twice: once against the stored mask and once against the tight bounding rectangle of the same mask. The tight rectangle is the most conservative axis-aligned bounding-box rule, because it is the smallest rectangle that contains the full target mask. Mask regrading reproduces the recorded server verdict on $99.5$--$100.0\%$ of rollouts in every cell; the remaining differences come from one-pixel rounding at the mask boundary. The comparison therefore changes only the grading rule while holding the clicks fixed.

\begin{table}[htb]
\centering
\small
\caption{How much of a target's bounding rectangle belongs to the target.
Mask and rectangle areas are averaged over the $200$ test puzzles of each
type. The final column gives the accuracy of a click drawn uniformly inside the rectangle under the mask verifier.}
\label{tab:rect_area}
\begin{tabular}{@{}lrrrrr@{}}
\toprule
Type & mask/image & rect/image & mask/rect (mean) & mask/rect (min--max) & uniform-in-rect \\
\midrule
Geometry Click & 0.8\% & 1.1\% & 0.78 & 0.57--0.97 & 78.3 \\
Pick Area      & 38.8\% & 65.1\% & 0.62 & 0.37--0.89 & 61.6 \\
\bottomrule
\end{tabular}
\end{table}

\begin{table}[htb]
\centering
\small
\caption{The same clicks evaluated under mask and bounding-rectangle grading.}
\label{tab:rect_regrade}
\setlength{\tabcolsep}{5pt}
\begin{tabular}{@{}lr rrr rrr@{}}
\toprule
& & \multicolumn{3}{c}{Geometry Click}
& \multicolumn{3}{c}{Pick Area} \\
\cmidrule(lr){3-5}\cmidrule(lr){6-8}
System & $n$ & Verifier & Rect & $\Delta$
       & Verifier & Rect & $\Delta$ \\
\midrule
Qwen3.5-9B (untrained)
& 200 & 0.0 & 0.0 & $+0.0$
& 35.5 & 71.5 & $+36.0$ \\

UI-TARS-1.5-7B
& 200 & 85.5 & 90.5 & $+5.0$
& 33.0 & 77.5 & $+44.5$ \\

SFT
& 1000 & 68.2 & 76.2 & $+8.0$
& 59.8 & 81.6 & $+21.8$ \\

RL
& 1000 & 68.2 & 73.4 & $+5.2$
& 60.6 & 82.3 & $+21.7$ \\
\midrule
\emph{uniform click in rect}
& & 78.3 & 100.0 & &
61.6 & 100.0 & \\
\bottomrule
\end{tabular}
\\[3pt]
{\scriptsize
\emph{Verifier}: grading against the stored pixel mask.
\emph{Rect}: grading the same clicks against the tight bounding rectangle.
$\Delta$: the resulting accuracy inflation.
$n$: number of graded rollouts.}
\end{table}

\paragraph{Reading.}
Table~\ref{tab:rect_area} shows why the two types behave differently.
A Geometry Click target is a small glyph whose bounding rectangle covers only $1.1\%$ of the image, so rectangle grading mainly admits near misses. A Pick Area target is a much larger irregular region, and its rectangle covers $65.1\%$ of the image, allowing substantially more background to be accepted.

Table~\ref{tab:rect_regrade} shows the effect on measured accuracy.
On Geometry Click, rectangle grading increases accuracy by $5.0$--$8.0$
points across UI-TARS and the two trained policies, while the untrained
backbone scores zero under both rules. On Pick Area, the increase is $21.7$--$21.8$ points for the trained policies, $36.0$ points for the untrained backbone, and $44.5$ points for UI-TARS. In particular, UI-TARS rises from $33.0$ under the verifier to $77.5$ under rectangle grading, even though its verifier score is below the $61.6$ obtained by a click sampled uniformly inside the rectangle. Bounding-box grading therefore changes the apparent quality of grounding rather than simply adding a constant offset.

\begin{figure}[htb]
\centering
\begin{subfigure}[t]{0.56\textwidth}
\centering
\includegraphics[width=\textwidth]{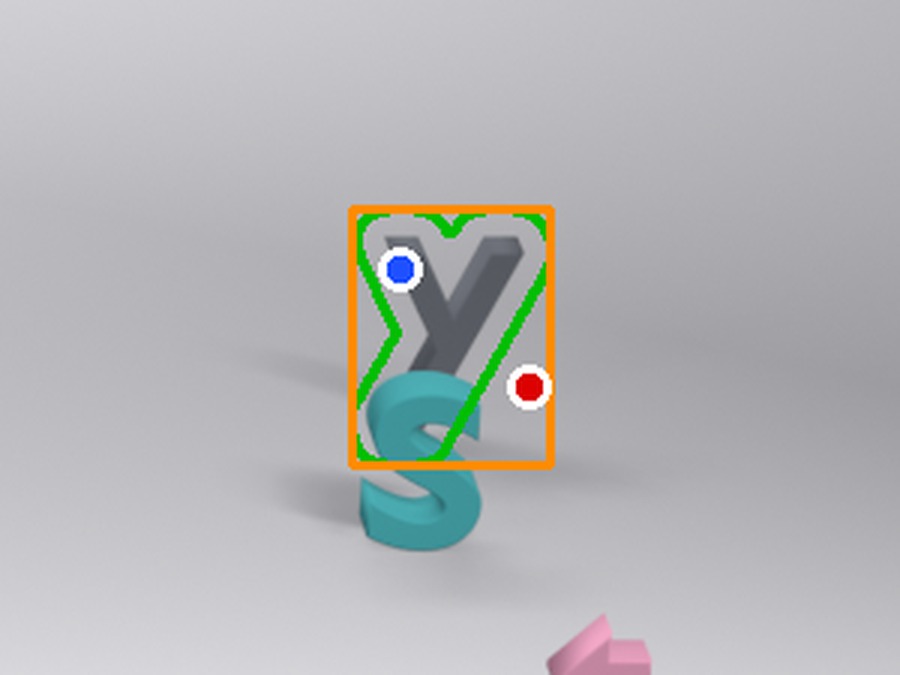}
\caption{Geometry Click image196, "click on the letter y".
UI-TARS lands inside the rectangle, $11$\,px off the mask.}
\label{fig:case_rect_A}
\end{subfigure}\hfill
\begin{subfigure}[t]{0.42\textwidth}
\centering
\includegraphics[width=\textwidth]{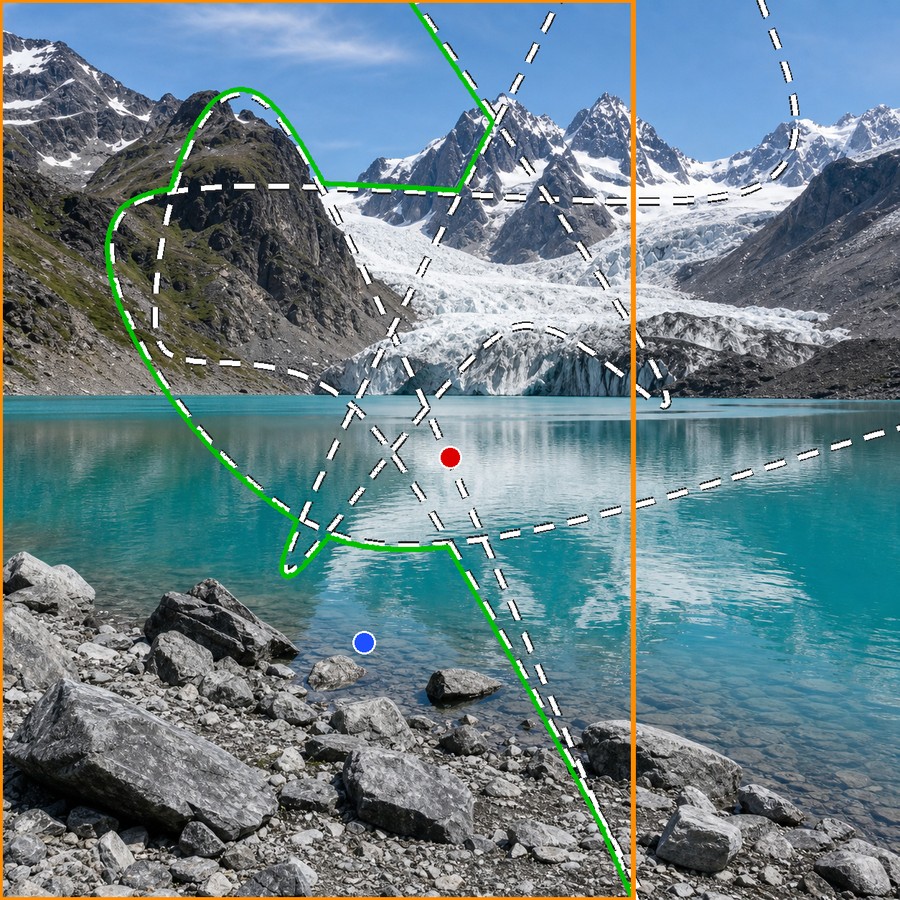}
\caption{Pick Area image2307. UI-TARS lands inside the rectangle
but outside the target region.}
\label{fig:case_rect_B}
\end{subfigure}
\caption{Two UI-TARS clicks accepted by bounding-rectangle grading but
rejected by the pixel-mask verifier. \textbf{CaptchaAgent} (RL) click on the same puzzle is shown in blue. Green denotes the verifier mask and orange its tight bounding rectangle.}
\label{fig:case_rect}
\end{figure}

Figure~\ref{fig:case_rect} shows two examples. The Geometry Click miss
(Fig.~\ref{fig:case_rect_A}) is a genuine near miss: the UI-TARS click
lands a few pixels away from the target glyph but remains inside its bounding rectangle. Rectangle grading would therefore count the click as correct even though it misses the target under the page's verifier. The Pick Area example (Fig.~\ref{fig:case_rect_B}) shows a larger grounding error: the click lands inside the bounding rectangle but outside the target region. These examples illustrate how rectangle grading can accept both small localization errors and clicks on background enclosed by an irregular target's bounding box.

\subsection{Where \textbf{CaptchaAgent} Succeeds and UI-TARS Fails}
\label{app:cases:modes}

\paragraph{Selection.}
We examine four types with different interaction requirements: Hold Button and Slide Puzzle from the real-time interaction mode, Rotation Match from arrow-cycle, and Bingo from multi-click. \textbf{CaptchaAgent} reaches $100.0$, $43.9$, $94.5$, and $56.3$ Pass@1 on these four types, respectively, while UI-TARS reaches $0.0$, $2.5$, $3.5$, and $3.5$. Hold Button additionally exposes an action-space limitation because UI-TARS does not provide a timed hold action.

\begin{figure}[t]
  \centering
{\renewcommand{\arraystretch}{0}%
  \begin{tabular}[t]{@{}c@{}}
    \rotatebox{90}{\makebox[0.1052\textwidth][c]{\scriptsize Hold Button}}\\[4pt]
    \rotatebox{90}{\makebox[0.1677\textwidth][c]{\scriptsize Slide Puzzle}}\\[4pt]
    \rotatebox{90}{\makebox[0.2130\textwidth][c]{\scriptsize Rotation Match}}\\[4pt]
    \rotatebox{90}{\makebox[0.2099\textwidth][c]{\scriptsize Bingo}}
  \end{tabular}}\hfill
  \begin{subfigure}[t]{0.30\textwidth}
    \centering
    \includegraphics[width=\textwidth]{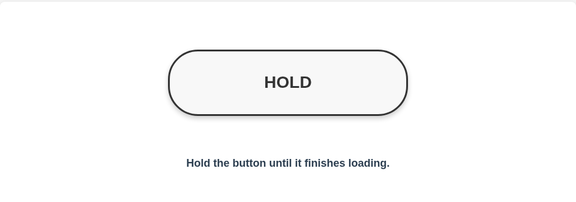}\\[3pt]
    \includegraphics[width=\textwidth]{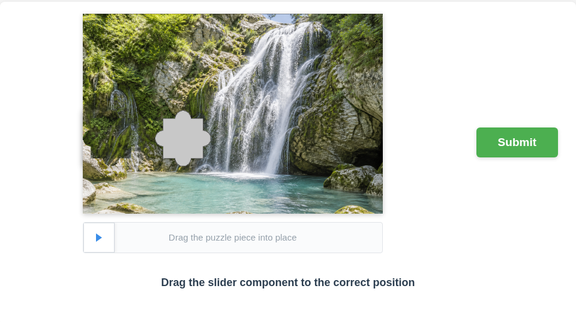}\\[3pt]
    \includegraphics[width=\textwidth]{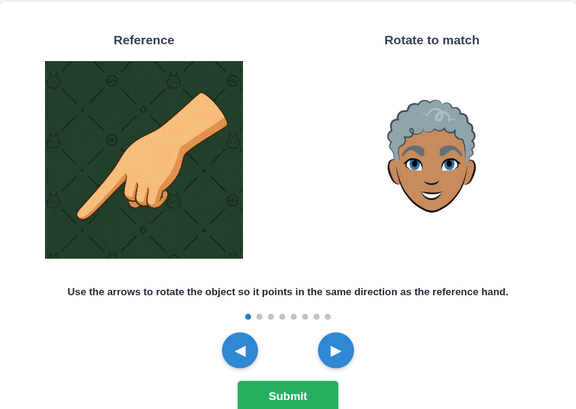}\\[3pt]
    \includegraphics[width=\textwidth]{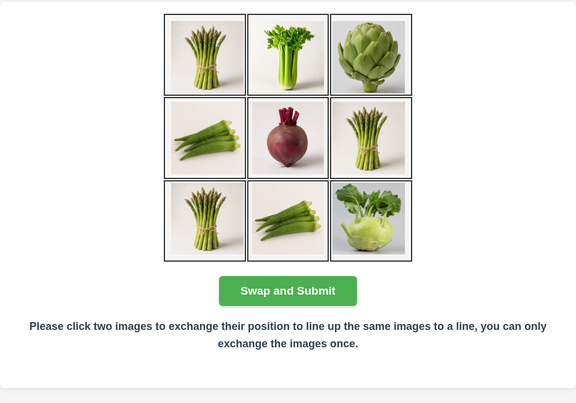}
    \caption{Initial page}
    \label{fig:case_modes_init}
  \end{subfigure}\hfill
  \begin{subfigure}[t]{0.30\textwidth}
    \centering
    \includegraphics[width=\textwidth]{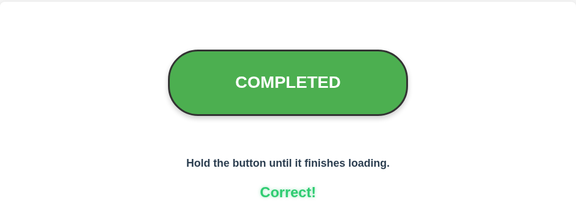}\\[3pt]
    \includegraphics[width=\textwidth]{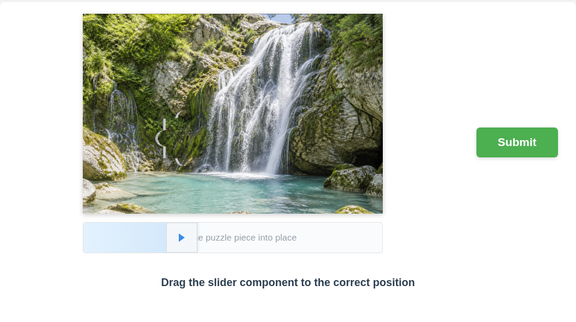}\\[3pt]
    \includegraphics[width=\textwidth]{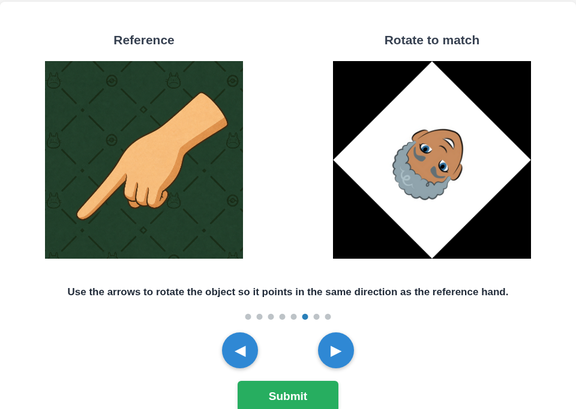}\\[3pt]
    \includegraphics[width=\textwidth]{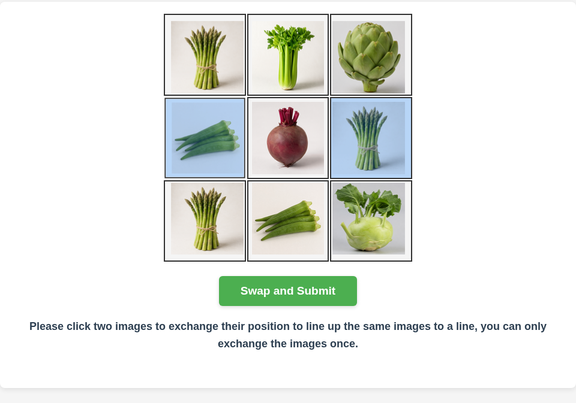}
    \caption{\textbf{CaptchaAgent}}
    \label{fig:case_modes_ours}
  \end{subfigure}\hfill
  \begin{subfigure}[t]{0.30\textwidth}
    \centering
    \includegraphics[width=\textwidth]{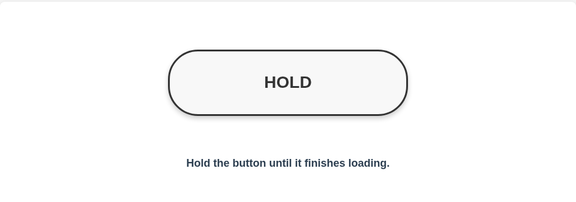}\\[3pt]
    \includegraphics[width=\textwidth]{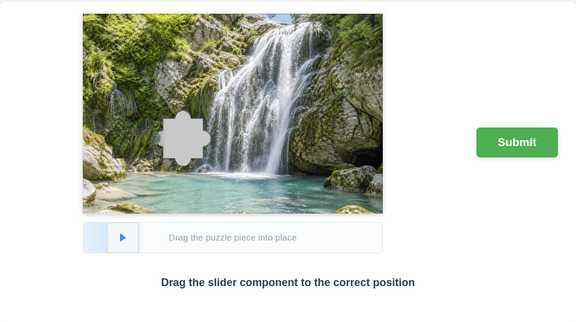}\\[3pt]
    \includegraphics[width=\textwidth]{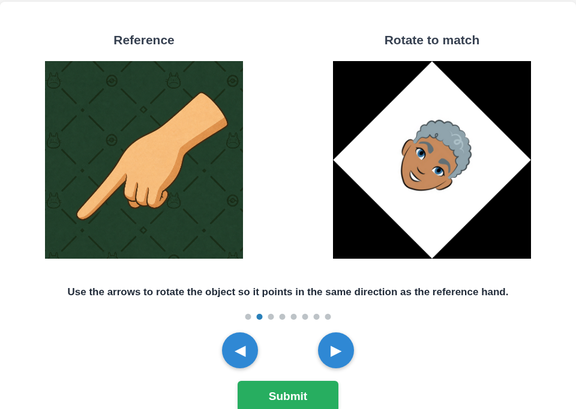}\\[3pt]
    \includegraphics[width=\textwidth]{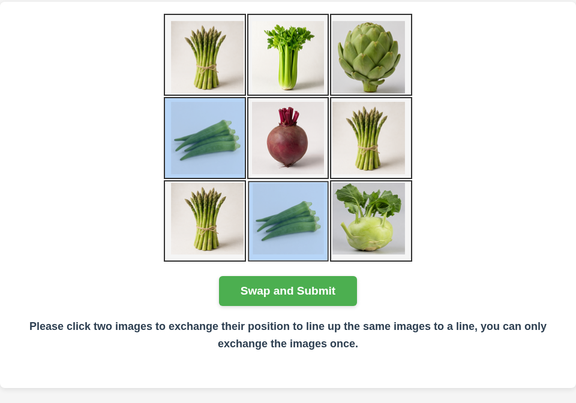}
    \caption{UI-TARS-1.5-7B}
    \label{fig:case_modes_tars}
  \end{subfigure}

  \caption{Four examples spanning real-time, arrow-cycle, and multi-click
  interactions. Column (a) shows the initial page; columns (b) and (c) show the last state reached by the \textbf{CaptchaAgent} and UI-TARS before submitting or stopping. Action traces and outcomes are listed in
  Table~\ref{tab:case_acts}.}
  \label{fig:case_modes}
\end{figure}

For each type, we show one test puzzle on which the \textbf{CaptchaAgent} succeeds and UI-TARS fails. The examples are Hold Button image2301, Slide Puzzle
slide2304, Rotation Match puzzle\_rotation\_105\_225, and Bingo bingo2303. Figure~\ref{fig:case_modes} shows the initial page and the final state reached by each system before submission or stopping, and
Table~\ref{tab:case_acts} lists the corresponding actions.

\begin{table}[t]
\centering
\scriptsize
\caption{Action traces behind Figure~\ref{fig:case_modes}. Coordinates are
viewport pixels; $\times k$ denotes repetition of the same action.
\emph{Step cap} means that the episode reaches the step limit without
submission.}
\label{tab:case_acts}
\setlength{\tabcolsep}{3pt}
\begin{tabular}{@{}llrl l@{}}
\toprule
Puzzle & System & Steps & Actions & Outcome \\
\midrule

\multirow{2}{*}{Hold Button}
 & \textbf{CaptchaAgent} & 1
 & hold(640,408,$\Delta t$)
 & correct \\
 & UI-TARS & 15
 & $14\times$click on and around the button;
   $1\times$drag
 & step cap \\
\midrule

\multirow{2}{*}{Slide Puzzle}
 & \textbf{CaptchaAgent} & 3
 & drag(325$\to$394);
   drag(394$\to$463) at $y{=}666$; submit
 & correct \\
 & UI-TARS & 12
 & $3\times$click on the handle;
   $9\times$drag of $6$--$18$\,px
 & no submission \\
\midrule

\multirow{2}{*}{Rotation Match}
 & \textbf{CaptchaAgent} & 6
 & click(720,847)$\times5$ (right arrow); submit
 & correct \\
 & UI-TARS & 7
 & left, left, right, left, right, right, right
 & no submission \\
\midrule

\multirow{2}{*}{Bingo}
 & \textbf{CaptchaAgent} & 3
 & click(779,505); click(501,505); submit
 & correct \\
 & UI-TARS & 13
 & $12\times$click toggling tiles across the grid; submit
 & wrong answer \\
\bottomrule
\end{tabular}
\end{table}

\paragraph{What the agents do.}
The four failures take different forms. On \textbf{Hold Button}, UI-TARS
repeatedly clicks on and around the button but cannot issue the required hold action and reaches the step limit. On \textbf{Slide Puzzle}, it uses repeated short drags of only $6$--$18$ pixels and stops before the piece reaches its slot, while \textbf{CaptchaAgent} completes the movement in two larger drags and submits. On \textbf{Rotation Match}, UI-TARS alternates left and right actions, stops one step from the target orientation, and terminates without submitting. The \textbf{CaptchaAgent} instead advances to the correct orientation and submits. On \textbf{Bingo}, UI-TARS repeatedly toggles tiles and ultimately submits an invalid swap, while the \textbf{CaptchaAgent} identifies and executes a valid two-tile swap.

These examples expose procedural and interface limitations rather than
failures of localization alone: the agent must hold for a duration, drag toward a target, track changes across successive states, and complete a valid multi-step interaction before submission. These behaviors are directly represented in the replay-verified trajectories, where
each intermediate screenshot and action is preserved. The comparison is
in-distribution for \textbf{CaptchaAgent} and zero-shot for UI-TARS, so it shows the behavior of an unadapted general GUI agent rather than what UI-TARS could achieve after task-specific training. The external evaluations in App.~\ref{app:ocw} provide the corresponding out-of-distribution comparison.

\section{Supervised Fine-Tuning Configuration}
\label{app:sftcfg}

Table~\ref{tab:sftcfg} gives the full configuration for \S\ref{sec:training}. LoRA adapters are applied only to the language model, while the vision encoder and vision--language aligner remain frozen. Visual inputs are therefore encoded using the pretrained vision components throughout training. We report the final checkpoint after the full training schedule rather than selecting the best epoch on the validation set.

Per-turn expansion converts the $12{,}000$ training puzzles into $37{,}621$ training samples, corresponding to an average of $3.1$ supervised turns per puzzle.

\begin{table}[h]
\centering
\footnotesize
\setlength{\tabcolsep}{10pt}
\caption{Supervised fine-tuning configuration.}
\label{tab:sftcfg}
\begin{tabular}{@{}ll@{}}
\toprule
\textbf{Configuration} & \textbf{Value} \\
\midrule
\emph{Model} & \\
Base model          & Qwen3.5-9B \\
Adapters            & rank-$64$ LoRA, $\alpha=128$, dropout $0.05$ \\
LoRA target modules & $q$, $k$, $v$, $o$, gate, up, down \\
Frozen              & vision encoder, vision--language aligner \\
\midrule
\emph{Data} & \\
Training puzzles       & $12{,}000$ ($600$ per type) \\
Training samples       & $37{,}621$ \\
Maximum context length & $32{,}768$ tokens \\
\midrule
\emph{Optimization} & \\
Optimizer            & AdamW \\
Learning rate        & $10^{-4}$, cosine decay, $5\%$ warmup \\
Weight decay         & $0.01$ \\
Gradient clipping    & $1.0$ \\
Epochs               & $3$ ($1{,}764$ steps) \\
Effective batch size & $64$ \\
Precision            & bf16, ZeRO-3 with optimizer offload \\
Hardware             & $4\times$ A100 40GB \\
\bottomrule
\end{tabular}
\end{table}

\section{Reinforcement Learning Details}
\label{app:rl}

\paragraph{Task pool.}
We construct the GRPO training pool at the puzzle level rather than the task level. Candidate puzzles are evaluated with $5$ rollouts under the supervised policy, and puzzles solved in some but not all rollouts are retained because they are more likely to produce within-group reward variance. Puzzles that are always solved or always failed provide little or no GRPO learning signal. Puzzle-level mining is important because a task-level average can hide both
easy and difficult instances within the same type.  We therefore retain
trainable puzzles from all $20$ types rather than pruning entire types based on their mean success rate.

Coverage across all task types also limits drift. The KL penalty is evaluated only on sampled trajectories, so a type absent from the pool receives no direct KL constraint on its own trajectories while the shared model parameters are updated. The resulting GRPO training pool contains $2{,}643$ puzzles across all $20$ task types.

\paragraph{Reward.}
Let
\[
p=\mathrm{clip}(\mathrm{progress},0,1), \qquad
f=\mathrm{clip}\left(
\frac{\mathrm{turns}}{\mathrm{max\ turns}},0,1
\right),
\]
where $p$ is task-specific partial credit and $f$ is the fraction of the turn
budget used. The final reward is
\[
r=\max\left(0,\;r_{\mathrm{raw}}-w_{\mathrm{turn}}f\right),
\]
with
\[
r_{\mathrm{raw}}=
\begin{cases}
1,
& \text{if submitted and correct},\\[3pt]
w_{\mathrm{prog}}p^\gamma+
w_{\mathrm{sub}}\mathbf{1}[\mathrm{acted}\wedge p\ge\tau],
& \text{if submitted and incorrect},\\[3pt]
w_{\mathrm{prog}}p^\gamma(1-\lambda f),
& \text{if not submitted}.
\end{cases}
\]
We use
$w_{\mathrm{prog}}=0.3$,
$w_{\mathrm{sub}}=0.05$,
$\gamma=1$,
$\tau=0.5$,
$\lambda=0.5$, and
$w_{\mathrm{turn}}=0.03$.

The reward addresses two shortcuts observed during training. First, an
ungated submission bonus can encourage a cheap action followed immediately by
submission. We therefore require sufficient progress before granting the
submission bonus. Second, without a cost for long unsuccessful trajectories,
the policy can continue probing until the turn limit instead of committing to
an answer. The turn-dependent terms penalize this behavior. Flooring the final
reward at zero prevents differences in episode length among zero-progress
rollouts from creating artificial within-group variance.

\paragraph{Optimization and compute.}
Training is strictly on-policy. Each optimization step samples $8$ puzzles
with $16$ rollouts per puzzle, giving $128$ trajectories, and the same batch is
used for the update. A step takes approximately $6$--$10$ hours, dominated by
multi-turn interaction with the browser environment. Table~\ref{tab:rlcfg}
gives the full configuration.

\begin{table}[h]
\centering
\footnotesize
\setlength{\tabcolsep}{10pt}
\caption{Reinforcement learning configuration.}
\label{tab:rlcfg}
\begin{tabular}{@{}ll@{}}
\toprule
\textbf{Configuration} & \textbf{Value} \\
\midrule
\emph{Model} & \\
Base model  & supervised checkpoint from \S\ref{sec:training} \\
Trainable   & full language-model weights \\
Frozen      & vision encoder, vision--language aligner \\
Precision   & bf16, FSDP with parameter and optimizer offload \\
\midrule
\emph{Algorithm} & \\
Advantage estimator & GRPO \\
Group size          & $16$ rollouts per puzzle \\
Batch size          & $8$ puzzles per step \\
Optimizer           & AdamW \\
Learning rate       & $10^{-6}$, constant, no warmup \\
Entropy coefficient & $0$ \\
Clip ratio (high)   & $0.2$ \\
KL                  & $0.005$ against the frozen supervised reference \\
Epochs              & $1$ \\
Checkpoints         & one per step \\
\midrule
\emph{Rollout} & \\
Engine                 & SGLang, asynchronous \\
Maximum prompt length  & $3{,}072$ tokens \\
Maximum response length& $12{,}288$ tokens \\
Maximum context length & $32{,}768$ tokens \\
Temperature            & $1.0$ \\
Turn cap               & $15$ \\
Parallel calls         & $1$ \\
Memory fraction        & $0.38$ \\
\midrule
Hardware & $1\times$H100 and $3\times$RTX PRO 6000 ($96$\,GB) \\
\bottomrule
\end{tabular}
\end{table}

\paragraph{Checkpoint evaluation.}
After each training step, a quick evaluation on the validation split monitors training and shortlists checkpoints. Shortlisted checkpoints are then evaluated on all $200$ validation puzzles per type with $5$ rollouts per puzzle, using the same agent interface and verifier as the main evaluation. We select the checkpoint with the highest validation Pass@1 and evaluate it once on the test
split. All reported RL test results come from this single test evaluation.

\end{document}